\documentclass[11pt]{article}

\usepackage[preprint]{acl}

\usepackage{times}
\usepackage{latexsym}
\usepackage[T1]{fontenc}
\usepackage[utf8]{inputenc}
\usepackage{microtype}
\usepackage{multirow}
\usepackage{graphicx}

\usepackage{tablefootnote}
\usepackage{booktabs}
\usepackage{tcolorbox}
\usepackage{listings}
\usepackage{amsmath}
\usepackage{algorithm}
\usepackage{algorithmic}
\usepackage{wrapfig}
\usepackage{caption}
\usepackage{subcaption}
\usepackage{hyperref}
\usepackage{url}

\definecolor{darkblue}{rgb}{0, 0, 0.5}
\definecolor{darkgreen}{RGB}{0,120,0}
\definecolor{darkred}{RGB}{160,0,0}

\hypersetup{colorlinks=true, citecolor=darkblue, linkcolor=darkblue, urlcolor=darkblue}

\title{On the Limits of Layer Pruning \\for Generative Reasoning in Large Language Models}

\author{
Safal Shrestha\thanks{Equal contribution}\textsuperscript{1} \quad
Anubhav Shrestha\footnotemark[1]\textsuperscript{2} \quad
Aadim Nepal\textsuperscript{2} \quad
Minwu Kim\textsuperscript{2,3} \quad
Keith Ross\thanks{Corresponding author: \texttt{keithwross@nyu.edu}}\textsuperscript{2}
\\
\textsuperscript{1}Center for Interdisciplinary Data Science and AI, New York University Abu Dhabi
\\
\textsuperscript{2}Department of Computer Science, New York University Abu Dhabi
\\
\textsuperscript{3}Seoul National University
}

\begin{document}
\maketitle

\begin{abstract}
Recent work has shown that layer pruning can effectively compress large language models (LLMs) while retaining strong performance on classification benchmarks, often with little or no finetuning. In contrast, generative reasoning tasks, such as GSM8K and HumanEval\textsuperscript{+}, exhibit substantially weaker recovery.
We show that beyond surface-level text degradation, pruning leads to a loss of key algorithmic capabilities, including arithmetic computation and balanced parenthesis generation. Under realistic post-training constraints, using a single 80GB GPU and without access to pretraining-scale data or compute, we evaluate a simple recovery strategy based on supervised finetuning with self-generated responses. This approach recovers up to 90\% of baseline performance on classification tasks, but recovery for generative reasoning remains limited.
We further find that this gap persists even under a favorable task-aligned recovery setting, where pruned models are fully finetuned on self-generated GSM8K responses, suggesting that the degradation is not merely due to generic instruction data or parameter-efficient tuning. As complementary evidence, we analyze a depth-pruned model trained with nearly 100B post-pruning tokens and find that deficits persist even on simple arithmetic tasks that do not require multi-step generation.
Overall, we characterize practical recovery limits of layer pruning for generative reasoning and provide guidance on when depth reduction is effective under constrained post-training regimes.%
\footnote{Code available at 
\url{https://github.com/safal312/on-the-limits-of-layer-pruning}
% \url{https://anonymous.4open.science/r/on-the-limits-of-layer-pruning-B355}
}
% Recent work has shown that layer pruning can effectively compress large language models (LLMs) while retaining strong performance on classification benchmarks, often with little or no finetuning. In contrast, generative reasoning tasks, such as GSM8K and HumanEval\textsuperscript{+}, exhibit substantially weaker recovery.
% We show that beyond surface-level text degradation, pruning leads to a loss of key algorithmic capabilities, including arithmetic computation and balanced parenthesis generation. Under realistic post-training constraints, without access to pretraining-scale data or compute, we evaluate a minimal recovery strategy based on supervised finetuning with self-generated responses. This approach recovers up to 90\% of baseline performance on classification tasks, but recovery for generative reasoning remains limited.
% Notably, even models finetuned on $\sim$100B tokens after pruning fail to recover their original reasoning performance, suggesting that such capabilities are not as easily restored. This limitation persists even on simple tasks such as arithmetic, which do not require multi-step generation.
% Overall, we characterize the limits of layer pruning for generative reasoning in practical recovery settings and provide guidance on when depth reduction is effective under constrained post-training regimes.%
% \footnote{Code available at \url{https://anonymous.4open.science/r/limit-of-layer-pruning-0167}}
\end{abstract}

\section{Introduction}

Large Language Models (LLMs) have achieved remarkable performance across a wide range of tasks, a success often attributed to their large parameter counts and extensive training data \citep{hoffmann2022training,yang2025qwen3,grattafiori2024llama}. However, the scale of modern LLMs raises significant concerns regarding efficiency and costs \citep{lecun1989optimal,wan2023efficient,song2024sleb}. These challenges have motivated a substantial body of work on model compression techniques aimed at reducing model size while preserving performance, including pruning at multiple granularities ranging from individual neurons to entire layers \citep{frantar2023sparsegptmassivelanguagemodels,sun2024simpleeffectivepruningapproach,muralidharan2024compact,sreenivas2024llm,song2024sleb,ashkboos2024slicegpt,ma2023llm,ling2024slimgpt}.

Among these approaches, layer pruning has emerged as a particularly appealing strategy. By removing entire transformer blocks of contemporary decoder-only models, layer pruning offers a simple method for reducing model depth, often requiring minimal or no additional finetuning \citep{song2024sleb,yang2024laco,lu2024reassessing,men2025shortgpt,kim2024shortened,chen2024streamlining}. This approach is further motivated by theoretical and empirical works suggesting redundancy across layers in LLMs \citep{sun2025curse,lad2024remarkable,men2025shortgpt}. Furthermore, layer pruning is largely orthogonal to other efficiency techniques such as quantization and sparsification, enabling it to be combined with complementary methods for additional computational savings \citep{song2024sleb,kim2024shortened}.

While this approach has achieved notable success in classification benchmarks, prior works show that it has proven far less effective for reasoning-intensive generative tasks like math and coding, which require the model to generate a multi-step chain of thought to arrive at the correct solution \citep{lu2024reassessing,chen2024streamlining,men2025shortgpt,yang2024laco,kim2024shortened,nepal2025layer}. Previous studies have largely attributed the failure of layer pruning on generative tasks to the importance of deeper layers for ``reasoning,'' without explicitly characterizing how layer removal degrades model behavior \citep{wang2025fewer,song2025demystifying}. Moreover, existing methods that partially recover generative performance typically require access to large-scale data (in billions of tokens) and compute, which can be impractical \citep{muralidharan2024compact,sreenivas2024llm}. These limitations motivate a closer analysis of pruning-induced failure modes and an examination of how much recovery is achievable under realistic post-training constraints (e.g., limited to a single $\sim$80GB GPU and moderate-scale data). Rather than proposing a new pruning algorithm, our goal is to characterize the limits of layer pruning in practical settings for generative reasoning and to identify regimes where it remains viable.

In this paper, we make the following contributions:

\begin{itemize}
    \item \textbf{We establish a controlled and realistic evaluation setting for post-pruning recovery.} Using self-generated responses (SGR) as a simple and strong baseline, we isolate the extent to which pruned models can recover capabilities under practical constraints (single 80GB GPU, no access to pretraining-scale data).
    
    \item \textbf{We systematically characterize the recovery gap induced by layer pruning for generative reasoning under realistic post-training constraints.} While prior work reports weaker performance on generative tasks, we show that this gap persists even with SGR across multiple model families, despite strong improvements on classification benchmarks.

    \item \textbf{We identify concrete failure modes underlying this degradation.} Beyond surface-level text degeneration, we show that pruning disrupts core algorithmic capabilities, including arithmetic computation and syntactic structure (e.g., parenthesis tracking), which are essential for multi-step reasoning.

    \item \textbf{We examine whether the generative reasoning gap persists under favorable recovery settings.} Using task-aligned full finetuning on self-generated GSM8K responses, we find that pruned models remain well below the unpruned baseline. As complementary evidence, we analyze Minitron, a depth-pruned model trained with nearly 100B post-pruning tokens, and find that reasoning and arithmetic deficits persist.
\end{itemize}

Overall, our results clarify when and why layer pruning succeeds or fails, and provide practical guidance for its use in settings where preserving generative reasoning ability is a priority.

\section{Layer Pruning and Recovery}

Layer pruning has been widely studied as a simple and effective compression strategy for LLMs \citep{men2025shortgpt,kim2024shortened,song2024sleb,yang2024laco,chen2024streamlining}. In this work, we study a standard post-pruning pipeline under compute-constrained settings: (1) remove $N$ layers according to a fixed strategy, and (2) recover performance through post-training. We consider two commonly used strategies for selecting layers: Block Influence (BI) and Reverse Order \citep{men2025shortgpt,muralidharan2024compact,wang2025fewer,lu2024reassessing}. 
BI removes layers whose output activations exhibit high cosine similarity with their inputs, as measured on a calibration dataset (e.g., C4 \cite{dodge2021documenting}), while Reverse Order prunes layers from the tail end of the model. Experiments in this paper are primarily conducted on Gemma2-2B-It, Llama-3.1-8B-It, Qwen-2.5-7B-Instruct, and Mistral-7B-Instruct-v0.3.

\paragraph{Post-Training with SGR.}
Following pruning, performance is typically recovered via supervised finetuning on high-quality prompt--response datasets \citep{chen2024streamlining,ma2023llm,gromov2024unreasonable,lu2024reassessing,wang2025fewer,xia2023sheared}. Prior work has also explored ``self-distillation'' approaches that refine or rewrite dataset responses using the base model \citep{thangarasa2025self}. It can be considered as a form of knowledge distillation where the unpruned model is the teacher and pruned model is the student \cite{hinton2015distilling}.

In this work, we adopt a simplified form of self-distillation in which we discard dataset responses entirely and instead use self-generated responses (SGR) on the prompts using the unpruned model. The resulting prompt--response pairs are then used to finetune the pruned model via QLoRA. This formulation removes dependence on reference outputs and isolates the extent to which the pruned model can recover capabilities from its own generated supervision. We use this approach as a simple and strong baseline for studying post-pruning recovery. Experiments are conducted on Alpaca-cleaned \citep{taori_alpaca_2023} and Dolci datasets \citep{olmo2025olmo}.

\begin{figure*}[t]
  \begin{center}
    \centerline{\includegraphics[width=0.8\textwidth]{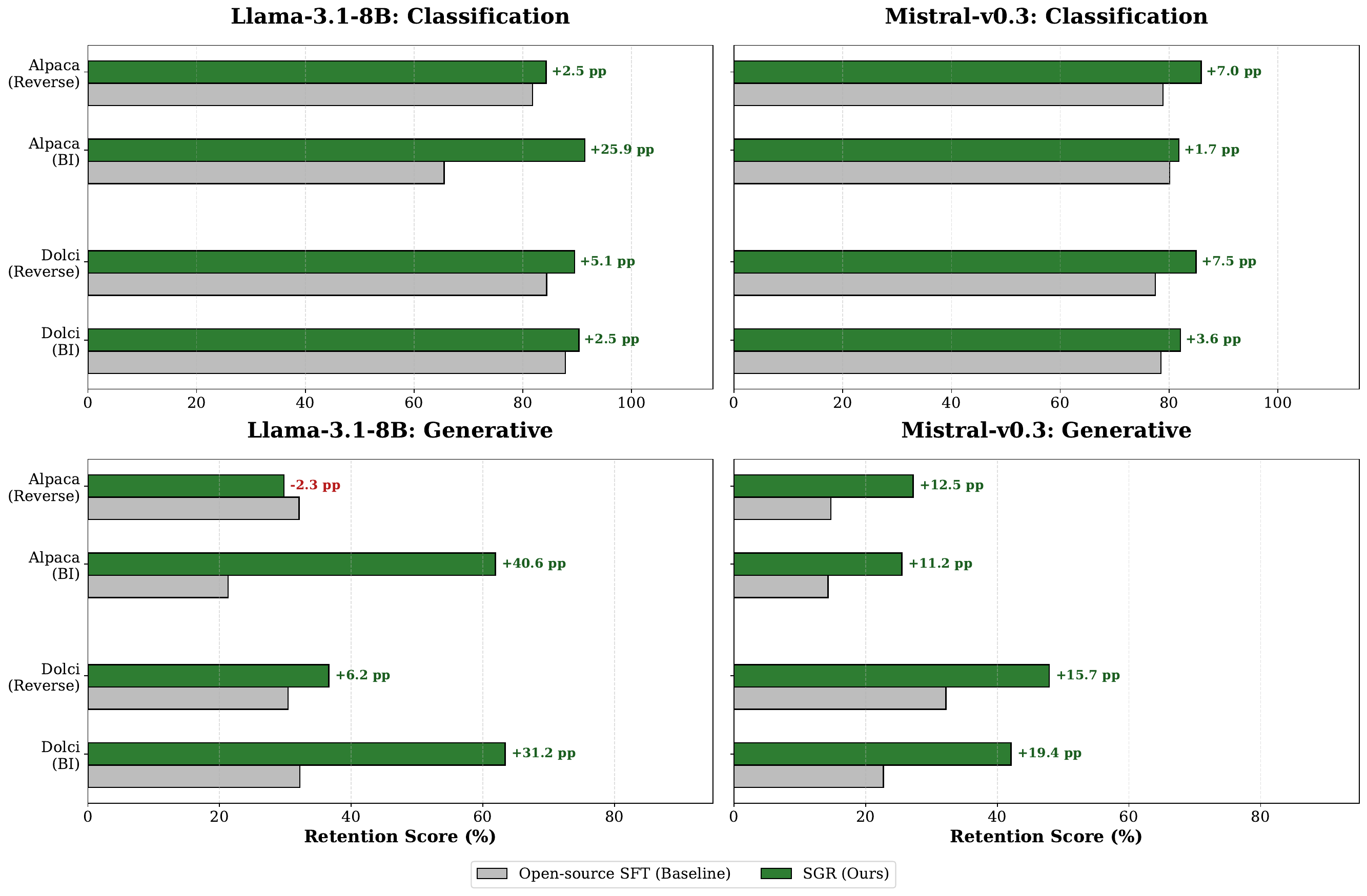}}
    \caption{
      Comparison of performance retention (normalized to baseline) between SFT with open source data versus SGR (our method), with BI and Reverse pruning metrics at 25\% pruning ratio. Full results in ~\ref{sec:self_generated_full}.
    }
    \label{fig:sgr-figure}
  \end{center}
\end{figure*}

\paragraph{Benchmarks.}
Following standard literature, in this paper, we evaluate on the following classification benchmarks: HellaSwag \citep{zellers2019hellaswag}, PIQA \citep{bisk2020piqa}, MMLU \citep{hendrycks2020measuring}, WinoGrande \citep{sakaguchi2021winogrande}, OpenBookQA \citep{mihaylov2018can}, ARC-E \citep{clark2018think}, and ARC-C \citep{clark2018think}. All classification results are computed using the \texttt{lm-evaluation-harness} framework \citep{gao2021framework}. We also consider a diverse set of generative benchmarks spanning multiple domains, including GSM8K for mathematical reasoning \citep{cobbe2021training}, HumanEval\textsuperscript{+} and MBPP\textsuperscript{+} for code generation \citep{liu2023your}, and XSUM for summarization \citep{narayan2018don}. While in classification benchmarks, evaluation reduces to scoring a fixed set of candidate outputs via log-likelihood comparison, in generative benchmarks, we require the model to produce a sequence of tokens to arrive at a valid solution, often involving multi-step reasoning or structured generation. These tasks place substantially different demands on the model than classification benchmarks.

This pipeline allows us to perform pruning and training with QLoRA \citep{dettmers2023qlora} on a single 80GB GPU, making it suitable for resource-constrained settings. Full experimental details are provided in Appendix~\ref{all-finetuning-appendix}. Although we use QLoRA for efficiency, we perform extensive comparisons showing that it achieves performance comparable to LoRA and even full finetuning (Table~\ref{tab:pruning_external_benchmarks_all}, Figure~\ref{fig:ft_vs_qlora}, Section \ref{upper-bound}).

\subsection{Results}
\label{classification-result-section}

As shown in Figure~\ref{fig:sgr-figure}, supervised finetuning with self-generated responses (SGR) consistently yields substantially stronger improvements than standard finetuning on fixed open-source datasets, in line with \citet{thangarasa2025self}. In particular, we observe that direct supervised finetuning on such datasets provides limited recovery after pruning, motivating the use of SGR as a stronger baseline for evaluating recoverability.

These gains are particularly pronounced for classification tasks. For example, for LLaMA pruned using the Block Influence (BI) metric, SGR training on Alpaca achieves an average retention of 91.4\%, corresponding to a +25.9 percentage point improvement over finetuning on Alpaca alone. SGR also improves performance on generative tasks. For instance, the same LLaMA model pruned with BI achieves a +31.2 percentage point gain when finetuned on Dolci SGR relative to finetuning on Dolci directly. However, a clear disparity remains: while classification performance typically retains close to or above 80\%, generative performance remains below 65\% in most cases. We observe similar trends across other model families, including Gemma and Qwen (see Appendix~\ref{sec:self_generated_full}). We further test whether reinforcement-learning finetuning on top of SGR-recovered models can close this gap, but find only limited additional recovery (Appendix~\ref{rlvr-recovery-appendix}).

Prior work has also observed that layer pruning can disproportionately harm generative reasoning compared to classification; our results are intended to evaluate this gap under a controlled SGR-based recovery setting rather than claim the disparity itself as new. Overall, these results demonstrate that while stronger supervision such as SGR can substantially improve recovery, especially for classification, a significant gap persists between classification and generative performance. For example, while the same LLaMA model retains nearly 90\% accuracy on classification benchmarks, its retention on generative benchmarks is only 63.4\% (Figure~\ref{fig:sgr-figure}). Although some degradation is expected due to the increased difficulty of generative reasoning tasks, these findings suggest that such capabilities remain harder to recover under the post-pruning recovery settings we evaluate.
We examine whether milder depth reduction offers a more favorable
trade-off. As shown in Appendix~\ref{throughput-ratios}, removing only a small
number of layers yields modest throughput gains while retaining much of the
original generative performance, whereas more aggressive pruning leads to
disproportionately larger reasoning degradation. \textbf{\emph{If preserving generative reasoning is a primary objective, aggressive layer pruning may be impractical.}}
% Overall, these results demonstrate that while stronger supervision such as SGR can substantially improve recovery, especially for classification, a significant gap persists between classification and generative performance. For example, while the same LLaMA model retains nearly 90\% accuracy on classification benchmarks, its retention on generative benchmarks is only 63.4\% (Figure~\ref{fig:sgr-figure}). Although some degradation is expected due to the increased difficulty of generative reasoning tasks, these findings suggest that such capabilities are fundamentally harder to recover, even under stronger post-training strategies. \textbf{\emph{If preserving generative reasoning is a primary objective, aggressive layer pruning may be impractical.}}

\subsection{Task-Aligned Recovery on GSM8K}

\begin{figure}[htbp]
  \begin{center}
    \centerline{\includegraphics[width=\columnwidth]{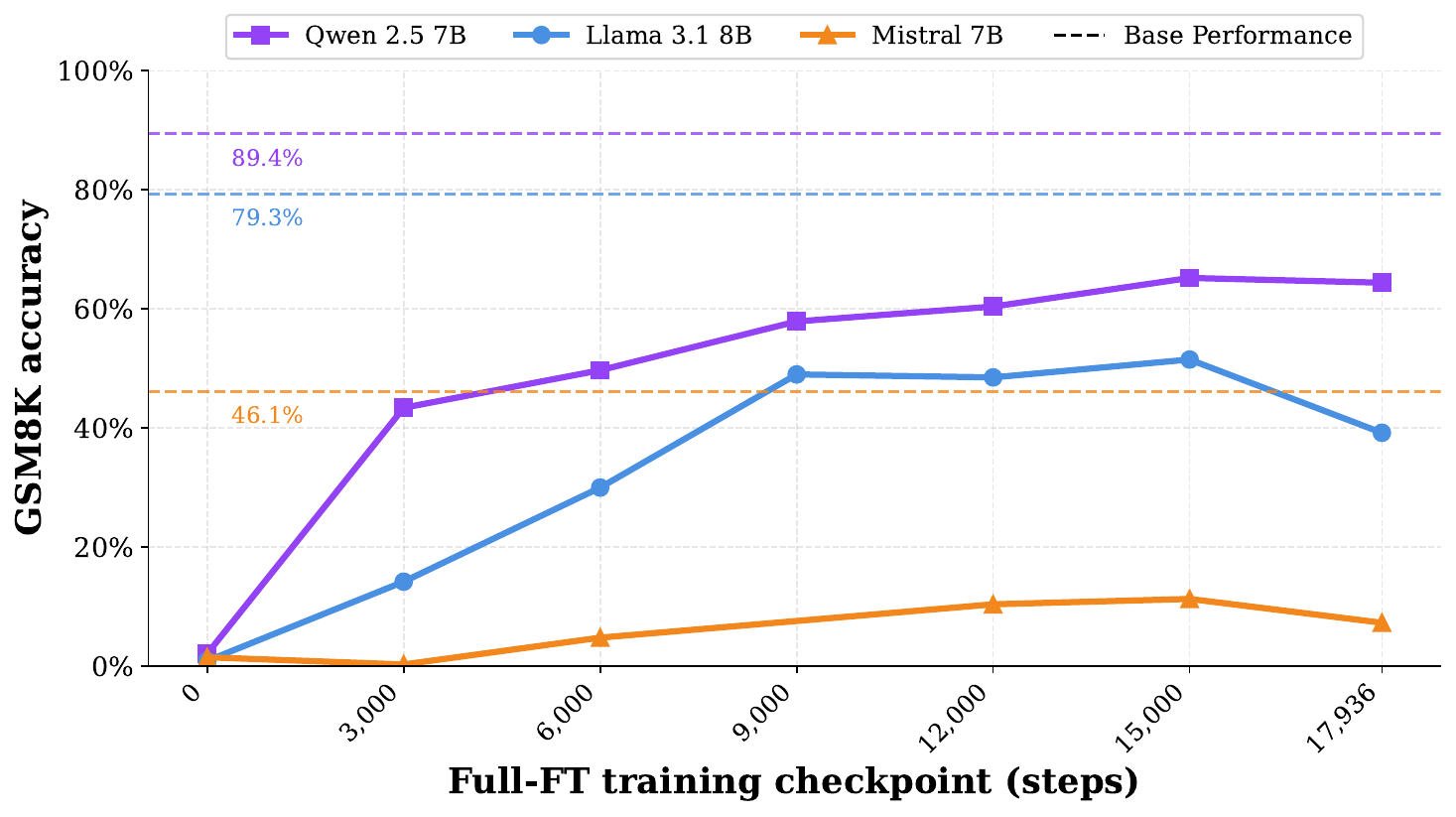}}
    \caption{
      Full Finetuning with GSM8K train set with SGR and evaluation on GSM8K test.
    }
    \label{fig:qlora_vs_sft_paper}
  \end{center}
\end{figure}
In addition, to test whether the recovery gap persists under a favorable task-aligned recovery setting, we conduct a stronger recovery ablation on GSM8K using the SGR method. We full-finetune pruned Qwen, LLaMA, and Mistral models on self-generated responses from the corresponding unpruned models using the GSM8K training split, and evaluate on GSM8K. This setting is more favorable than our main recovery experiments because the recovery data and evaluation task are aligned, and recovery uses full-parameter finetuning rather than QLoRA.

Despite these favorable conditions, Figure~\ref{fig:qlora_vs_sft_paper} shows that GSM8K performance remains substantially below the corresponding unpruned baselines across model families. LLaMA improves from 0.9\% to a peak of 51.5\%, but remains below its 79.3\% unpruned baseline; Mistral reaches only 11.3\%, compared to its 46.1\% baseline. Qwen shows the same qualitative trend, improving from 2.1\% to 65.2\% while remaining below its 89.4\% baseline. These results suggest that the recovery gap is not simply an artifact of generic instruction data or QLoRA-based recovery: even task-aligned full finetuning does not restore the original reasoning performance. More details are provided in Appendix~\ref{upper-bound}.

\subsection{Case Study on Minitron}
\label{minitron-llama}

As complementary external evidence, we also evaluate the publicly released 
\texttt{Llama-3.1-Minitron-4B-Depth} checkpoint~\citep{sreenivas2024llm}\footnote{The publicly released checkpoint corresponds to the model after post-pruning distillation ($\sim$94B tokens) but before the additional alignment stages (math/code SFT, instruction tuning, and preference optimization) used for the final results in \citet{sreenivas2024llm}. We therefore apply SGR finetuning before evaluation; see Appendix~\ref{app:minitron_alignment}.}, 
which combines 50\% depth pruning with approximately 94B tokens of 
post-pruning distillation. Since Minitron differs from our controlled setup 
in pruning ratio, layer-selection procedure, training data, and post-training 
pipeline, we treat it only as suggestive evidence. After SGR finetuning on 
Dolci, the model achieves 23.9\% on GSM8K and 15.34\% on MBPP\textsuperscript{+}, 
corresponding to 49.2\% and 34.3\% retention, respectively. This suggests that 
substantial post-pruning distillation alone does not necessarily restore 
reasoning performance, consistent with our controlled recovery results. We 
provide details and caveats in Appendix~\ref{app:minitron_alignment}.

\paragraph{Scope of the recovery study.}
Our primary setting is deliberately resource-constrained: we study post-pruning recovery with access to a single 80GB GPU, moderate-scale instruction data, and no pretraining-scale data or compute. We therefore use QLoRA as our main recovery method because it is practical under this budget. The task-aligned full-parameter finetuning experiments (Section~2.2), RLVR experiments (Appendix~A.8), and Minitron case study (Section~2.3) serve as complementary stress tests that relax different aspects of this setting. Together, they show that the observed recovery gap cannot be attributed solely to QLoRA, generic instruction data, or the single-GPU constraint. We do not, however, interpret these experiments as establishing that recovery is impossible with layer pruning.

\section{Layer-by-Layer Pruning for Generative Tasks}
\label{layer-by-layer-pruning}

\begin{figure*}[t]
  \begin{center}
    \centerline{\includegraphics[width=\textwidth]{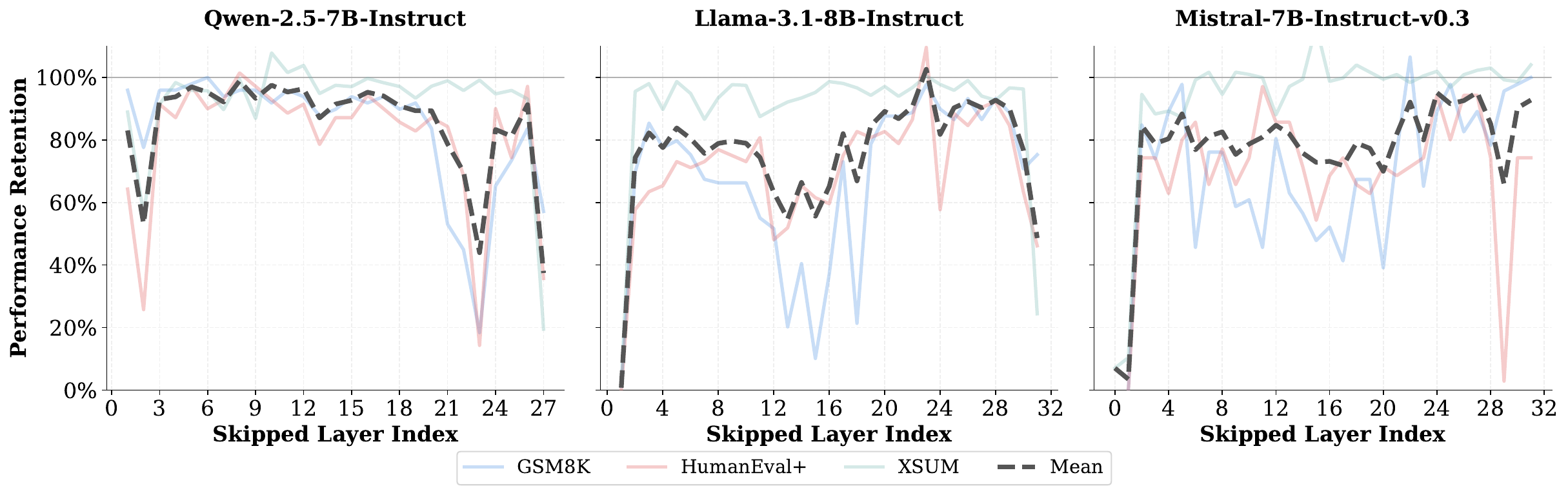}}
    \caption{
      Effect of removing a single layer on model performance across generative benchmarks. Reasoning-intensive tasks such as GSM8K and HumanEval\textsuperscript{+} exhibit severe performance degradation at specific layers, while XSUM remains comparatively robust except for layers whose removal induces general text degeneration. (We generally skip layer 0 for its poor results.)
    }
    \label{fig:multi-model-ablation}
  \end{center}
\end{figure*}

To investigate the mechanisms underlying the persistent recovery gap, we analyze the effects of removing individual transformer layers. Specifically, we perform single-layer pruning by removing one transformer layer at a time and evaluating the resulting model on a diverse set of generative benchmarks: GSM8K, HumanEval\textsuperscript{+}, and XSUM. Experiments are conducted on three models from distinct families: Qwen-2.5-7B-Instruct \citep{qwen2025qwen25technicalreport}, Llama-3.1-8B-Instruct \citep{grattafiori2024llama}, and Mistral-7B-Instruct-v0.3 \citep{jiang2023mistral7b}.

Figure~\ref{fig:multi-model-ablation} summarizes the results. In a few cases, certain layers appear redundant; for example, early layers in Qwen and some deeper layers in Llama can be removed with minimal effect. Overall, across models, even single-layer removal can substantially impact performance. Pruning certain layers can even lead to sharp drops on GSM8K and HumanEval\textsuperscript{+}. While prior work has noted performance degradation in mathematical reasoning under layer pruning \citep{chen2024streamlining,wang2025fewer,nepal2025layer}, we additionally observe similar sensitivity in code generation, with the locations of the sharp drops varying across model families and tasks, as seen in Qwen versus Mistral. Task-specific effects are also pronounced: reasoning-intensive tasks such as mathematics and coding are far more sensitive to depth reduction, whereas summarization tasks like XSUM remain largely stable. {\bf {\em These results indicate that layer pruning exhibits strong model- and task-dependent effects, contrasting with its relative robustness on classification benchmarks.}}

Although earlier studies have suggested that pruning affects reasoning \citep{gromov2024unreasonable,men2025shortgpt}, detailed characterization of the resulting errors remains limited. In the following sections, we analyze a subset of key failure modes that arise in generative reasoning after pruning.

\subsection{Text Degeneration}

\begin{figure}[b]
  \centering
  \includegraphics[width=1.0\columnwidth]{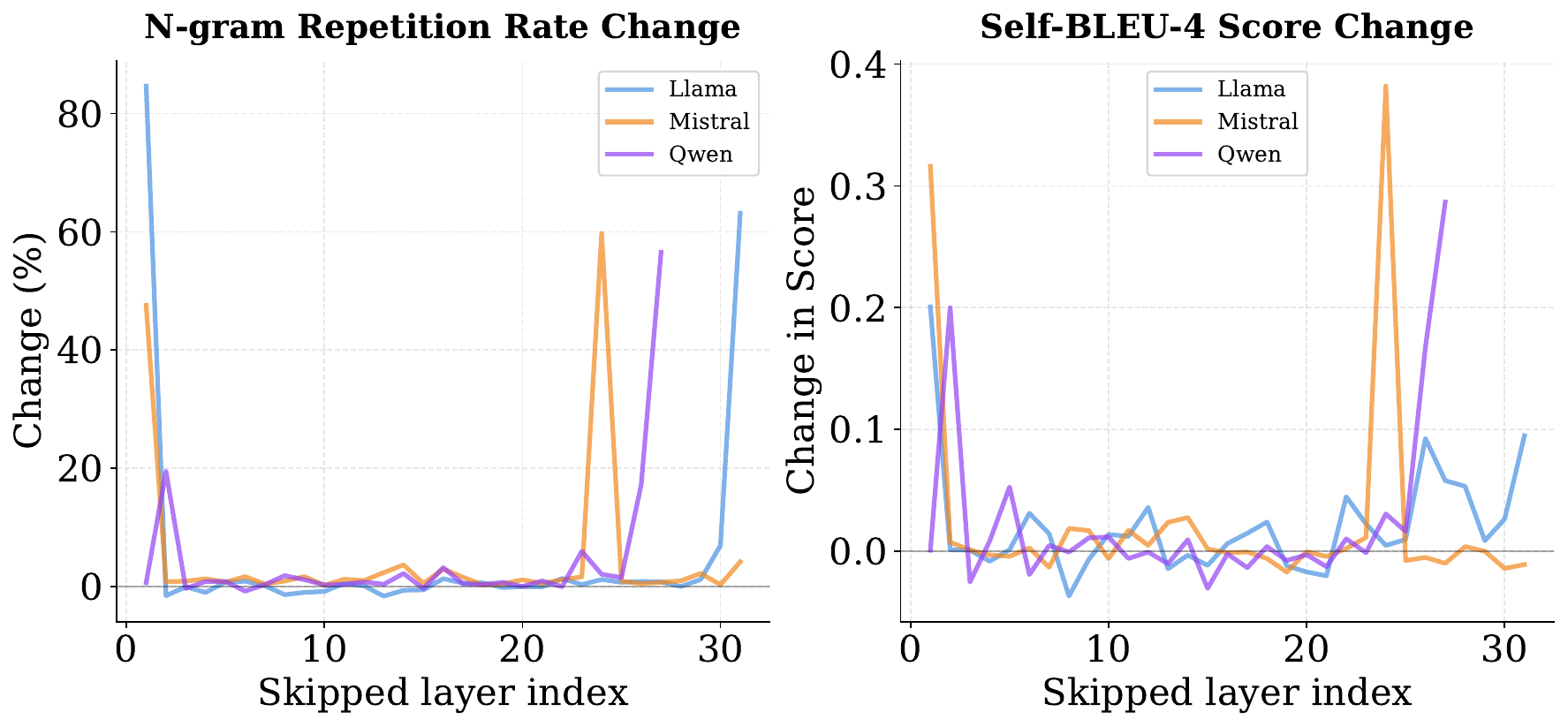}
  \caption{Text degeneration under single-layer pruning, measured using 4-gram repetition (left) and Self-BLEU4 averaged across responses and normalized relative to the baseline.}
  \label{fig:text-degeneration-main}
\end{figure}

Text degeneration \citep{holtzman2019curious} is a commonly observed failure mode in pruned language models and can hinder instruction following and coherent generation. We quantify degeneration using two complementary metrics computed with 4-grams: \emph{4-gram repetition} and \emph{Self-BLEU4}, where higher values indicate increased repetition and reduced diversity \citep{holtzman2019curious}. We additionally report the average number of generated tokens per prompt in Appendix~\ref{tokens-text-generation}.

As shown in Figure~\ref{fig:text-degeneration-main}, layer pruning often amplifies degenerative behaviors. In agreement with earlier findings \citep{men2025shortgpt}, we observe that both early and late layers are particularly important for maintaining stable text generation. In some cases, elevated repetition metrics align with sharp performance drops, such as layer 2 in Qwen.

Prior work has primarily attributed pruning-induced performance degradation to looping and repetitive outputs \citep{wang2025fewer}. However, degeneration alone does not fully account for failures in generative reasoning tasks. Near the points of sharp drops in Qwen and Llama on the math and coding tasks, text generation quality remains largely intact. Conversely, in Mistral, we observe a pronounced spike in degeneration metrics at layer 24 without a corresponding drop in task performance. Manual inspection reveals that removing this layer causes the model to continue rambling after producing a valid response. {\bf {\em Overall, these findings indicate that while text degeneration is a significant side effect of layer pruning, it is not a sufficient explanation for the loss of reasoning ability in generative tasks.}}

\begin{figure}[htbp]
  \centering
  \begin{subfigure}{\columnwidth}
    \centering
    \includegraphics[width=0.8\linewidth]{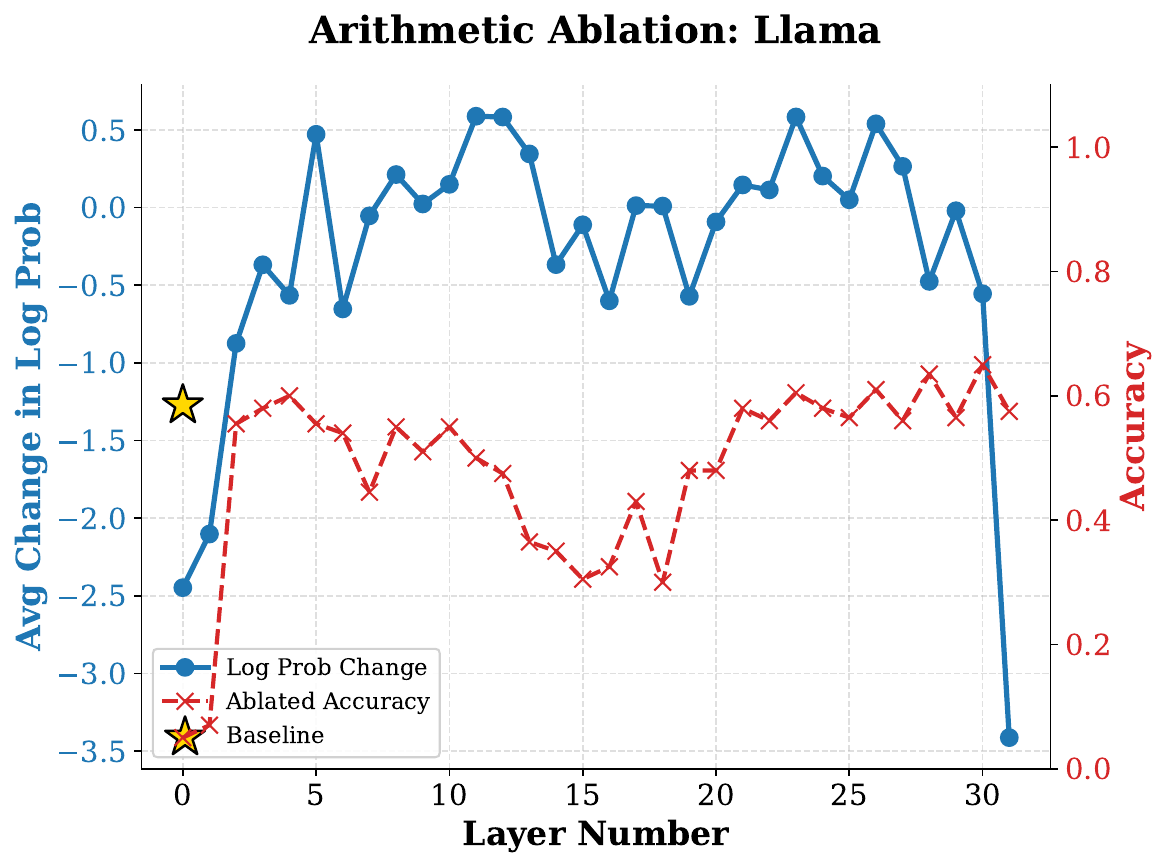}
    \caption{Effect of single-layer pruning on the arithmetic ability of Llama.}
    \label{fig:llama-arithmetic-error}
  \end{subfigure}

  \vspace{0.75em}

  \begin{subfigure}{\columnwidth}
    \centering
    \includegraphics[width=1.0\linewidth]{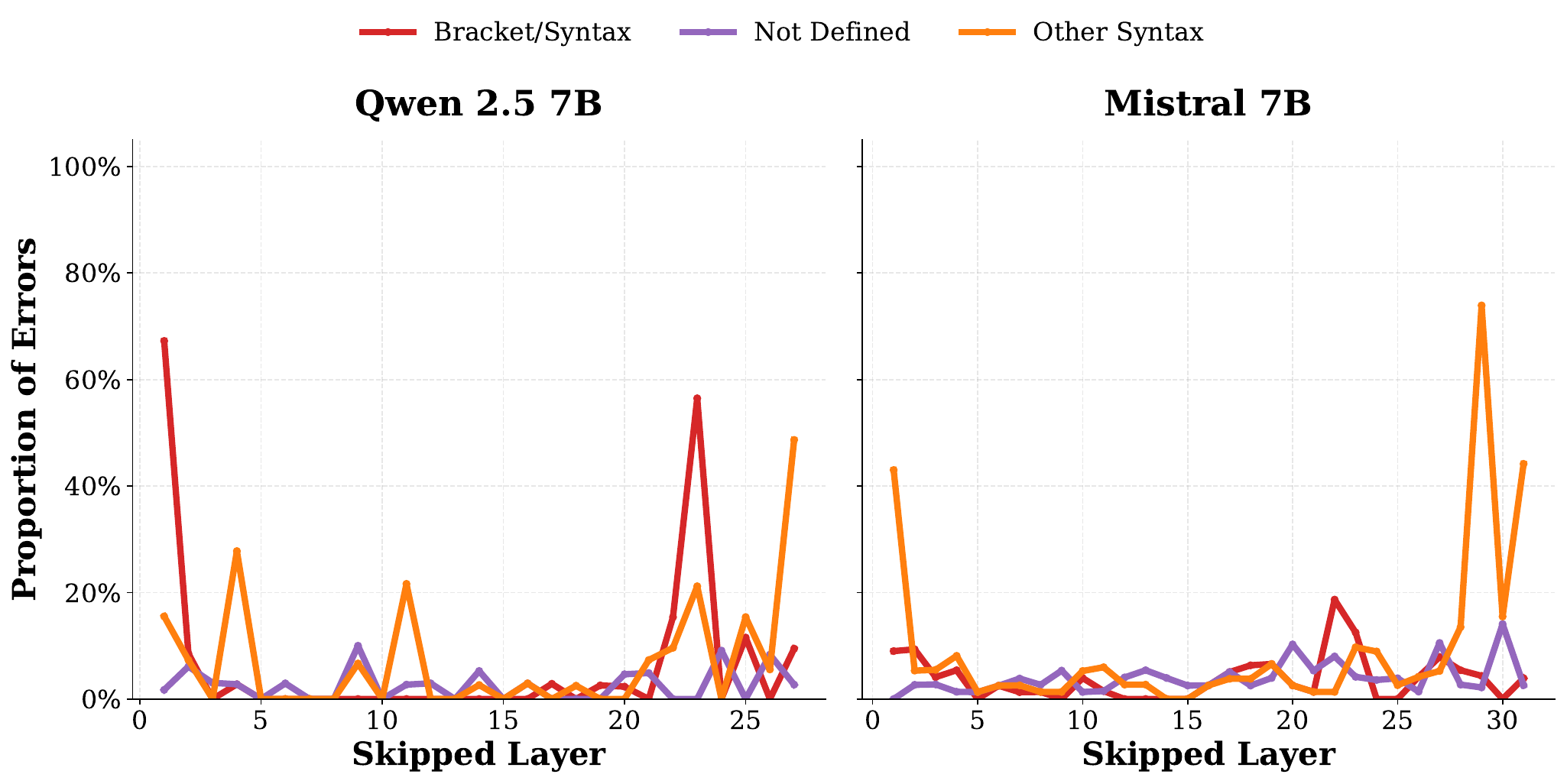}
    \caption{Distribution of syntactic error types induced by single-layer pruning.}
    \label{fig:syntax-error}
  \end{subfigure}

  \caption{Analysis of model performance and error types under single-layer pruning.}
  \label{fig:combined-analysis}
\end{figure}

% \begin{figure}[htbp]
%   \centering
%   \begin{subfigure}{0.40\columnwidth}
%     \centering
%     \includegraphics[width=\linewidth]{figures/llama_dual_axis_arithmetic_ablation.pdf}
%     \caption{Effect of single-layer pruning on the arithmetic ability of Llama.}
%     \label{fig:llama-arithmetic-error}
%   \end{subfigure}
%   \hfill
%   \begin{subfigure}{0.57\columnwidth}
%     \centering
%     \includegraphics[width=\linewidth]{figures/coding_degeneration_prop.pdf}
%     \caption{Distribution of syntactic error types induced by single-layer pruning.}
%     \label{fig:syntax-error}
%   \end{subfigure}
%   \caption{Analysis of Model Performance and Error Types under Single-Layer Pruning.}
%   \label{fig:combined-analysis}
% \end{figure}

\subsection{Degradation of Arithmetic}
\label{arithmetic-error}

Beyond high-level reasoning behaviors, solving mathematical word problems requires reliable execution of basic arithmetic operations. We find that pruning leads to a pronounced degradation of arithmetic ability, even on elementary computations. In particular, removing deeper layers in Qwen and mid-depth layers in Llama results in frequent arithmetic failures during qualitative inspection, where models fail to correctly perform simple calculations (see Appendix~\ref{arithmetic-mistake-example}).

Since standard generative evaluations can obscure true abilities due to auxiliary demands such as multi-token generation~\citep{schaeffer2023emergent,hu2024auxiliary,song2025demystifying}, to isolate arithmetic competence, we design a controlled evaluation that probes the model's ability to produce the \emph{first answer token} in simple arithmetic prompts. Given inputs of the form \texttt{``Question: What is (7 + 5) - 6? Answer:''}, we measure (i) the change in logprob of the correct answer token relative to the unpruned baseline, and (ii) top-1 accuracy, i.e., whether the correct token is assigned the highest logprob. Full experimental details are provided in Appendix~\ref{arithmetic-ablation-details}.

Figure~\ref{fig:llama-arithmetic-error} reports results over 200 arithmetic problems. We observe substantial drops in both logprob and accuracy after pruning layers near the middle region, despite the absence of any generation requirement in this task. {\bf {\em This demonstrates that layer pruning induces a degradation of arithmetic capability itself, rather than merely degrading Chain of Thought generation.}} Moreover, the correspondence between arithmetic failures and performance drops on GSM8K (Figure~\ref{fig:multi-model-ablation}) suggests that a significant fraction of mathematical reasoning errors stem directly from impaired arithmetic abilities. Results for Qwen and Mistral are reported in Appendix~\ref{arithmetic-ablation-details}.

\subsection{Degradation of Parenthesis Tracking}
\label{parenthesis-tracking}

Figure~\ref{fig:multi-model-ablation} shows that layer pruning can cause substantial performance drops on coding benchmarks such as HumanEval\textsuperscript{+}. Similar to arithmetic in mathematical reasoning, correct syntax generation is a necessary prerequisite for code reasoning. {\bf \em We observe that removing specific layers significantly degrades the model's ability to maintain syntactic consistency, particularly in tracking and closing parentheses.} Representative examples are provided in Appendix~\ref{parenthesis-error}.

Unlike math benchmarks, code evaluation allows errors to be categorized based on execution feedback. Leveraging this property, Figure~\ref{fig:syntax-error} reports the prevalence of common syntactic failures under single-layer pruning for Qwen and Mistral, including \textit{unbalanced parentheses}, \textit{undefined variables}, and \textit{other invalid syntax}. Notably, pruning certain layers (e.g., layer 23 in Qwen) leads to a sharp increase in parenthesis-matching errors, indicating a severe loss in syntactic skills. In Mistral, we also observe spikes in auxiliary syntax failures, such as malformed markdown code blocks (e.g., \texttt{```python (code)```}), which can directly disrupt downstream evaluation pipelines.
\\
\newline
Taken together, these results indicate that layer pruning disrupts not only surface-level text generation but also, potentially, internal mechanisms responsible for important algorithmic capabilities, such as arithmetic execution and parenthesis tracking. Tasks such as classification or summarization are less dependent on specialized capabilities, which helps explain their relative robustness to depth reduction. Single-layer ablations are diagnostic rather than predictive. We do not claim that the effect of removing multiple layers is a linear composition of single-layer effects. Instead, single-layer experiments identify candidate failure modes induced by depth disruption. Section 4 then evaluates whether the same failures persist in realistic multi-layer pruning and recovery settings. 
% In contrast, the pronounced sensitivity observed under single-layer removal highlights why aggressive depth pruning poses a fundamental challenge for generative reasoning tasks, making performance retention without continued training almost impossible.

\section{Post-Recovery Analysis}

\begin{figure}[htbp]
  \centering
  \begin{subfigure}{\columnwidth}
    \centering
    \includegraphics[width=0.7\linewidth]{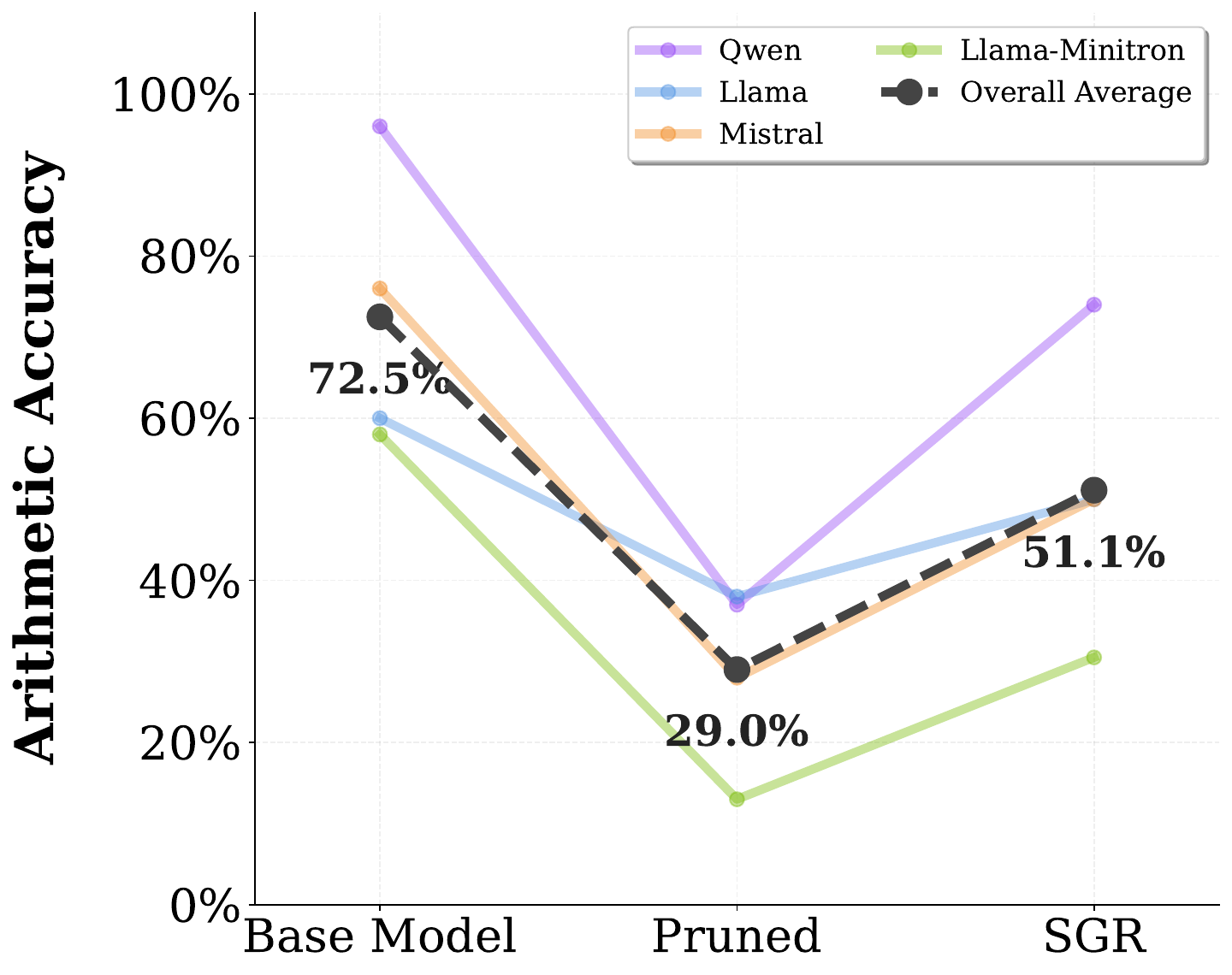}
    \caption{Arithmetic task accuracy for baseline, pruned, and finetuned models (50\% ratio for Minitron, 25\% for rest).}
    \label{fig:arithmetic-post-recovery}
  \end{subfigure}

  \vspace{0.75em}

  \begin{subfigure}{\columnwidth}
    \centering
    \includegraphics[width=0.7\linewidth]{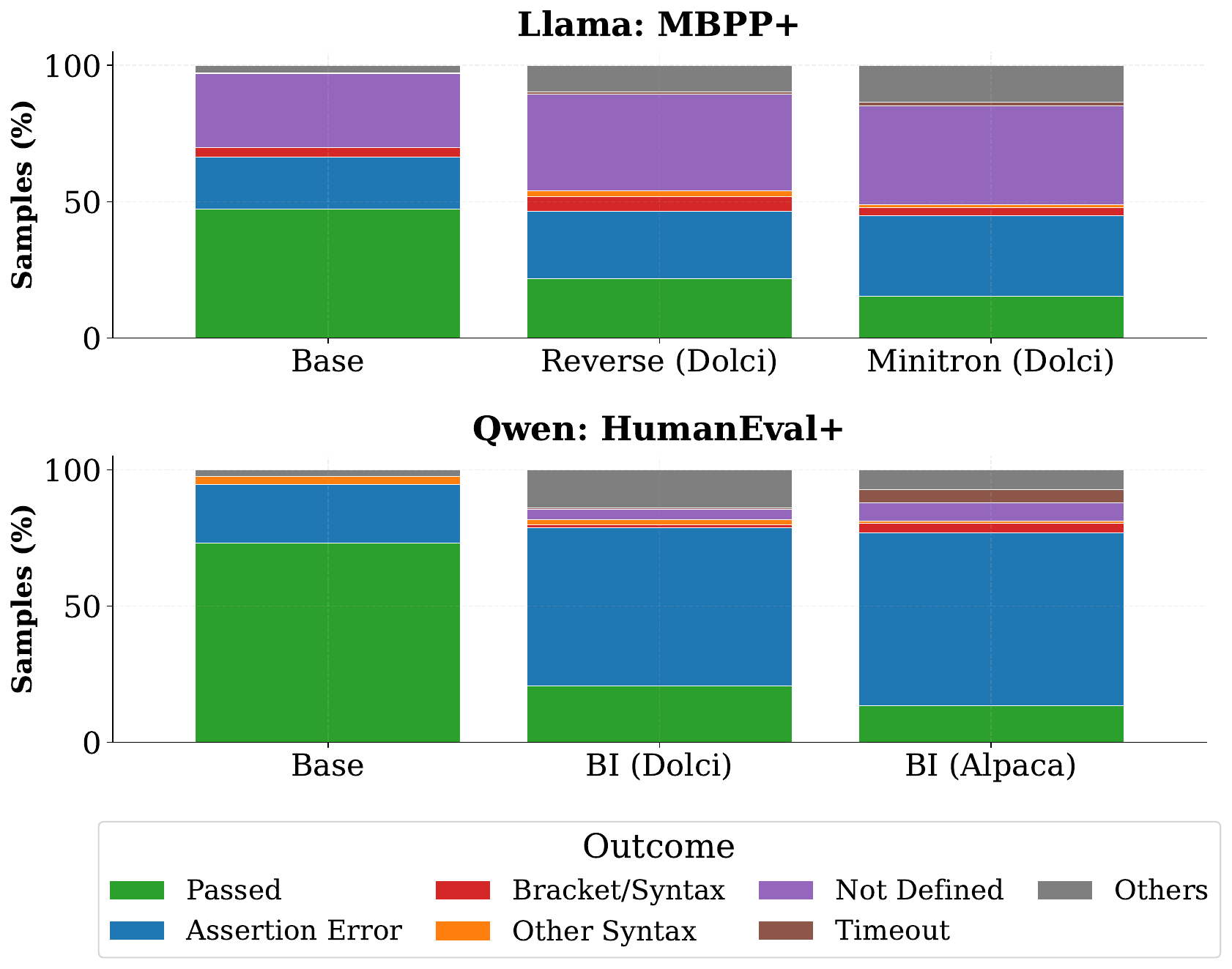}
    \caption{Code evaluation outcomes (MBPP\textsuperscript{+} and HumanEval\textsuperscript{+}) across model families.}
    \label{fig:coding-compression}
  \end{subfigure}

  \caption{Comparison of model performance under pruning and recovery.}
  \label{fig:combined-performance-analysis}
\end{figure}

% \begin{figure}[htbp]
%   \centering
%   \begin{subfigure}{0.48\columnwidth}
%     \centering
%     \includegraphics[width=\linewidth]{figures/arithmetic_average_trend.pdf}
%     \caption{Arithmetic task accuracy for baseline, pruned, and finetuned models (50\% ratio for Minitron, 25\% for rest).}
%     \label{fig:arithmetic-post-recovery}
%   \end{subfigure}
%   \hfill
%   \begin{subfigure}{0.48\columnwidth}
%     \centering
%     \includegraphics[width=\linewidth]{figures/compact_comparison.pdf}
%     \caption{Code evaluation outcomes (MBPP\textsuperscript{+} and HumanEval\textsuperscript{+}) across model families.}
%     \label{fig:coding-compression}
%   \end{subfigure}
%   \caption{Comparison of model performance across arithmetic and coding benchmarks under pruning and recovery.}
%   \label{fig:combined-performance-analysis}
% \end{figure}

\subsection{Arithmetic Ability}
\label{arithmetic-recovery}

Building on the analysis in Section~\ref{arithmetic-error}, we examine how arithmetic ability is affected by pruning and the extent to which it can be recovered through SGR finetuning. Figure~\ref{fig:arithmetic-post-recovery} reports results for three model families at a fixed pruning ratio of 25\%.

Across all models, pruning leads to a substantial drop in arithmetic accuracy, with average accuracy falling from 72.5\% before pruning to 29\% after pruning. Finetuning with SGR partially recovers this loss to 51.1\%, but performance remains well below that of the base model. Notably, this arithmetic task is considerably simpler than full generative mathematical reasoning, as it does not require multi-step reasoning or long-form generation. The persistent gap between base and finetuned pruned models, even on this minimal objective, indicates that \textbf{\emph{pruning degrades core computational capabilities in a manner that is difficult to recover.}} 

While some models, such as LLaMA, exhibit smaller absolute drops (e.g., from 60\% to 50\%), this residual gap represents a lower bound on the error introduced by pruning. For more complex reasoning tasks, which combine arithmetic with multi-step generation, this degradation is likely to be further amplified.

We further corroborate this trend using the Minitron model introduced in Section~\ref{minitron-llama}. The base model (LLaMA-3.1-8B) achieves 58\% accuracy on this task, which drops sharply to 13\% after pruning. Strikingly, even after nearly 100B tokens of post-pruning training followed by supervised finetuning with SGR, the model fails to recover simple arithmetic performance. This complements our controlled findings and suggests that the arithmetic degradation observed after pruning is not merely an artifact of small-scale recovery: even after large-scale post-pruning distillation, this publicly available depth-pruned checkpoint retains substantial deficits in basic arithmetic ability.
% This reinforces our earlier findings and highlights a fundamental limitation: even extensive post-training is insufficient to restore basic computational abilities once depth has been reduced.

\subsection{Syntax Ability}
\label{syntax-recovery}

As discussed in Section~\ref{parenthesis-tracking}, coding benchmarks are particularly sensitive to layer pruning, as they require strict syntactic correctness in addition to semantic reasoning. Figure~\ref{fig:coding-compression} summarizes post-recovery code evaluation results across model families. {\bf \em Even after finetuning, pruned models continue to have difficulty generating syntactically valid code.} This issue is more pronounced on MBPP\textsuperscript{+}, where code must be generated from natural language descriptions, compared to HumanEval\textsuperscript{+}, which provides a function signature as a prefix. Even the Minitron model, after significant training, fails to handle logical and syntactically sound code generation.

Overall, finetuning only partially restores syntactic abilities after pruning. Across models, we observe persistent errors such as undefined variables and unbalanced parentheses, consistent with the failure modes identified in Section~\ref{parenthesis-tracking}. The prevalence of invalid and logically incorrect code indicates that pruning impairs the model's ability to maintain syntactic and state consistency during generation, which are not fully recoverable through training.

Although for Qwen in HumanEval\textsuperscript{+}, we see higher rates of syntactically valid code, the fraction of functionally correct solutions remains substantially lower. This disparity highlights that syntactic validity alone does not imply recovery of coding competence. Analogous to arithmetic serving as a lower bound for mathematical reasoning, these results suggest that residual syntactic errors provide a lower bound on the degradation of coding ability, with deeper semantic reasoning likely affected to an even greater extent.

\section{Background \& Related Work}

\paragraph{Layer Pruning in LLMs.}
Recent work suggests that layers in LLMs, particularly deeper ones, exhibit redundancy, motivating layer pruning as an effective compression strategy \citep{gromov2024unreasonable,sun2025curse,men2025shortgpt,yin2023outlier,lad2024remarkable}. On classification benchmarks, layer pruning achieves strong results, with models retaining over 80\% of baseline accuracy even after removing 20--25\% of layers \citep{yang2024laco,men2025shortgpt,gromov2024unreasonable,song2024sleb}. However, these gains do not consistently extend to generative reasoning tasks such as GSM8K, which require multi-step generation \citep{chen2024streamlining,men2025shortgpt,gromov2024unreasonable,wang2025fewer}. This discrepancy suggests that layer redundancy is task-dependent and motivates studying not only whether performance drops, but which capabilities are disrupted by depth reduction.

\paragraph{Limitations for Generative Reasoning.}
Prior work largely attributes the degradation in generative performance to the importance of deeper layers for reasoning \citep{song2025demystifying,wang2025fewer}. Some studies highlight changes in evaluation or qualitative differences in chain-of-thought outputs after pruning \citep{song2025demystifying,wang2025fewer}, but a systematic analysis of the underlying failure modes and their persistence after recovery remains limited. To mitigate performance loss, existing approaches rely on continued pretraining \citep{gromov2024unreasonable,sreenivas2024llm,muralidharan2024compact,xia2023sheared} or architectural modifications \citep{chen2024streamlining}, often requiring large-scale data and compute. In more constrained settings, standard finetuning on open-source datasets has been commonly used, but yields limited recovery for generative tasks \citep{sreenivas2024llm,muralidharan2024compact,gromov2024unreasonable,wang2025fewer}. Our work focuses on this gap by studying post-pruning recovery under practical constraints and diagnosing persistent failures in arithmetic computation and structured generation.

\section{Conclusion}
In this work, we study layer pruning for generative reasoning under practical post-training constraints. Across SGR, task-aligned full-parameter finetuning, and RLVR, with Minitron providing complementary evidence, pruned models recover classification performance more readily than generative reasoning under the settings evaluated. Our diagnostic analyses link this gap to persistent losses in arithmetic and syntactic capabilities, rather than only surface-level generation quality.

These findings show that layer pruning is strongly task-dependent: moderate pruning may remain useful for classification and less reasoning-intensive tasks, whereas aggressive depth reduction can severely impair reasoning. Preserving these capabilities may therefore require involved multi-stage training or large scale compute, which is beyond the scope of resource-constrained setups.

\section*{Limitations}

Our study focuses primarily on practical post-pruning recovery regimes: moderate-scale instruction data, limited compute, and QLoRA-based finetuning. Although we include additional checks with task-aligned full finetuning, RLVR, and the Minitron case study, these do not constitute an exhaustive scaling study over all possible recovery methods. We also note that Minitron differs from our controlled experiments in pruning ratio, layer-selection strategy, training data, and post-training pipeline, so we treat it as complementary evidence rather than a clean causal comparison. Finally, our failure-mode analysis focuses on arithmetic and parenthesis/syntax tracking, which are important but not exhaustive aspects of generative reasoning. Future work could extend this analysis to broader forms of structured generation and targeted recovery. We also caution practitioners that an aggressively pruned model may appear largely intact on standard classification benchmarks while its performance on deployed reasoning tasks silently degrades; pruned models should therefore be validated directly on the generative tasks they are intended for, rather than on classification metrics alone.

\section*{Acknowledgements}
This work is submitted in part by the NYUAD Center for Interdisciplinary Data Science \& AI (CIDSAI), funded by Tamkeen under
the NYUAD Research Institute Award CG016 and the NYU Abu Dhabi Center for Artificial Intelligence and
Robotics, funded by Tamkeen under the Research Institute Award CG010. This research was carried out using the resources of the Center for Research Computing at New York University Abu Dhabi.

\bibliography{references}

@article{hoffmann2022training,
  title={Training compute-optimal large language models},
  author={Hoffmann, Jordan and Borgeaud, Sebastian and Mensch, Arthur and Buchatskaya, Elena and Cai, Trevor and Rutherford, Eliza and Casas, Diego de Las and Hendricks, Lisa Anne and Welbl, Johannes and Clark, Aidan and others},
  journal={arXiv preprint arXiv:2203.15556},
  year={2022}
}

@article{thangarasa2025self,
  title={Self-data distillation for recovering quality in pruned large language models},
  author={Thangarasa, Vithursan and Venkatesh, Ganesh and Lasby, Mike and Sinnadurai, Nish and Lie, Sean},
  journal={Proceedings of Machine Learning and Systems},
  volume={7},
  year={2025}
}

@inproceedings{dodge2021documenting,
  title={Documenting large webtext corpora: A case study on the colossal clean crawled corpus},
  author={Dodge, Jesse and Sap, Maarten and Marasovi{\'c}, Ana and Agnew, William and Ilharco, Gabriel and Groeneveld, Dirk and Mitchell, Margaret and Gardner, Matt},
  booktitle={Proceedings of the 2021 conference on empirical methods in natural language processing},
  pages={1286--1305},
  year={2021}
}

@article{yang2025qwen3,
  title={Qwen3 technical report},
  author={Yang, An and Li, Anfeng and Yang, Baosong and Zhang, Beichen and Hui, Binyuan and Zheng, Bo and Yu, Bowen and Gao, Chang and Huang, Chengen and Lv, Chenxu and others},
  journal={arXiv preprint arXiv:2505.09388},
  year={2025}
}

@article{grattafiori2024llama,
  title={The llama 3 herd of models},
  author={Grattafiori, Aaron and Dubey, Abhimanyu and Jauhri, Abhinav and Pandey, Abhinav and Kadian, Abhishek and Al-Dahle, Ahmad and Letman, Aiesha and Mathur, Akhil and Schelten, Alan and Vaughan, Alex and others},
  journal={arXiv preprint arXiv:2407.21783},
  year={2024}
}

@article{wan2023efficient,
  title={Efficient large language models: A survey},
  author={Wan, Zhongwei and Wang, Xin and Liu, Che and Alam, Samiul and Zheng, Yu and Liu, Jiachen and Qu, Zhongnan and Yan, Shen and Zhu, Yi and Zhang, Quanlu and others},
  journal={arXiv preprint arXiv:2312.03863},
  year={2023}
}

@article{frantar2023sparsegptmassivelanguagemodels,
      title={SparseGPT: Massive Language Models Can Be Accurately Pruned in One-Shot}, 
      author={Elias Frantar and Dan Alistarh},
      year={2023},
      journal={arXiv preprint arXiv:2301.00774},
}

@article{sun2024simpleeffectivepruningapproach,
      title={A Simple and Effective Pruning Approach for Large Language Models}, 
      author={Mingjie Sun and Zhuang Liu and Anna Bair and J. Zico Kolter},
      year={2024},
      journal={arXiv preprint arXiv:2306.11695},
}

@article{muralidharan2024compact,
  title={Compact language models via pruning and knowledge distillation},
  author={Muralidharan, Saurav and Turuvekere Sreenivas, Sharath and Joshi, Raviraj and Chochowski, Marcin and Patwary, Mostofa and Shoeybi, Mohammad and Catanzaro, Bryan and Kautz, Jan and Molchanov, Pavlo},
  journal={Advances in Neural Information Processing Systems},
  volume={37},
  pages={41076--41102},
  year={2024}
}

@article{sreenivas2024llm,
  title={Llm pruning and distillation in practice: The minitron approach},
  author={Sreenivas, Sharath Turuvekere and Muralidharan, Saurav and Joshi, Raviraj and Chochowski, Marcin and Mahabaleshwarkar, Ameya Sunil and Shen, Gerald and Zeng, Jiaqi and Chen, Zijia and Suhara, Yoshi and Diao, Shizhe and others},
  journal={arXiv preprint arXiv:2408.11796},
  year={2024}
}

@article{lecun1989optimal,
  title={Optimal brain damage},
  author={LeCun, Yann and Denker, John and Solla, Sara},
  journal={Advances in neural information processing systems},
  volume={2},
  year={1989}
}

@article{song2024sleb,
  title={Sleb: Streamlining llms through redundancy verification and elimination of transformer blocks},
  author={Song, Jiwon and Oh, Kyungseok and Kim, Taesu and Kim, Hyungjun and Kim, Yulhwa and Kim, Jae-Joon},
  journal={arXiv preprint arXiv:2402.09025},
  year={2024}
}

@article{ma2023llm,
  title={Llm-pruner: On the structural pruning of large language models},
  author={Ma, Xinyin and Fang, Gongfan and Wang, Xinchao},
  journal={Advances in neural information processing systems},
  volume={36},
  pages={21702--21720},
  year={2023}
}

@article{ashkboos2024slicegpt,
  title={Slicegpt: Compress large language models by deleting rows and columns},
  author={Ashkboos, Saleh and Croci, Maximilian L and Nascimento, Marcelo Gennari do and Hoefler, Torsten and Hensman, James},
  journal={arXiv preprint arXiv:2401.15024},
  year={2024}
}

@article{yang2024laco,
  title={Laco: Large language model pruning via layer collapse},
  author={Yang, Yifei and Cao, Zouying and Zhao, Hai},
  journal={arXiv preprint arXiv:2402.11187},
  year={2024}
}

@article{lu2024reassessing,
  title={Reassessing layer pruning in llms: New insights and methods},
  author={Lu, Yao and Cheng, Hao and Fang, Yujie and Wang, Zeyu and Wei, Jiaheng and Xu, Dongwei and Xuan, Qi and Yang, Xiaoniu and Zhu, Zhaowei},
  journal={arXiv preprint arXiv:2411.15558},
  year={2024}
}

@inproceedings{men2025shortgpt,
  title={Shortgpt: Layers in large language models are more redundant than you expect},
  author={Men, Xin and Xu, Mingyu and Zhang, Qingyu and Yuan, Qianhao and Wang, Bingning and Lin, Hongyu and Lu, Yaojie and Han, Xianpei and Chen, Weipeng},
  booktitle={Findings of the Association for Computational Linguistics: ACL 2025},
  pages={20192--20204},
  year={2025}
}

@article{kim2024shortened,
  title={Shortened llama: Depth pruning for large language models with comparison of retraining methods},
  author={Kim, Bo-Kyeong and Kim, Geonmin and Kim, Tae-Ho and Castells, Thibault and Choi, Shinkook and Shin, Junho and Song, Hyoung-Kyu},
  journal={arXiv preprint arXiv:2402.02834},
  year={2024}
}

@article{chen2024streamlining,
  title={Streamlining redundant layers to compress large language models},
  author={Chen, Xiaodong and Hu, Yuxuan and Zhang, Jing and Wang, Yanling and Li, Cuiping and Chen, Hong},
  journal={arXiv preprint arXiv:2403.19135},
  year={2024}
}

@article{ling2024slimgpt,
  title={Slimgpt: Layer-wise structured pruning for large language models},
  author={Ling, Gui and Wang, Ziyang and Liu, Qingwen},
  journal={Advances in Neural Information Processing Systems},
  volume={37},
  pages={107112--107137},
  year={2024}
}

@article{gromov2024unreasonable,
  title={The unreasonable ineffectiveness of the deeper layers},
  author={Gromov, Andrey and Tirumala, Kushal and Shapourian, Hassan and Glorioso, Paolo and Roberts, Daniel A},
  journal={arXiv preprint arXiv:2403.17887},
  year={2024}
}

@article{wang2025fewer,
  title={When Fewer Layers Break More Chains: Layer Pruning Harms Test-Time Scaling in LLMs},
  author={Wang, Keyu and Lyu, Tian and Su, Guinan and Geiping, Jonas and Yin, Lu and Canini, Marco and Liu, Shiwei},
  journal={arXiv preprint arXiv:2510.22228},
  year={2025}
}

@article{song2025demystifying,
  title={Demystifying the roles of llm layers in retrieval, knowledge, and reasoning},
  author={Song, Xinyuan and Wang, Keyu and Li, PengXiang and Yin, Lu and Liu, Shiwei},
  journal={arXiv preprint arXiv:2510.02091},
  year={2025}
}

@article{sun2025curse,
  title={The curse of depth in large language models},
  author={Sun, Wenfang and Song, Xinyuan and Li, Pengxiang and Yin, Lu and Zheng, Yefeng and Liu, Shiwei},
  journal={arXiv preprint arXiv:2502.05795},
  year={2025}
}

@article{yin2023outlier,
  title={Outlier weighed layerwise sparsity (owl): A missing secret sauce for pruning llms to high sparsity},
  author={Yin, Lu and Wu, You and Zhang, Zhenyu and Hsieh, Cheng-Yu and Wang, Yaqing and Jia, Yiling and Li, Gen and Jaiswal, Ajay and Pechenizkiy, Mykola and Liang, Yi and others},
  journal={arXiv preprint arXiv:2310.05175},
  year={2023}
}

@article{lad2024remarkable,
  title={The remarkable robustness of llms: Stages of inference?},
  author={Lad, Vedang and Lee, Jin Hwa and Gurnee, Wes and Tegmark, Max},
  journal={arXiv preprint arXiv:2406.19384},
  year={2024}
}

@article{xia2023sheared,
  title={Sheared llama: Accelerating language model pre-training via structured pruning},
  author={Xia, Mengzhou and Gao, Tianyu and Zeng, Zhiyuan and Chen, Danqi},
  journal={arXiv preprint arXiv:2310.06694},
  year={2023}
}

@inproceedings{dettmers2023case,
  title={The case for 4-bit precision: k-bit inference scaling laws},
  author={Dettmers, Tim and Zettlemoyer, Luke},
  booktitle={International Conference on Machine Learning},
  pages={7750--7774},
  year={2023},
  organization={PMLR}
}

@article{frantar2022gptq,
  title={Gptq: Accurate post-training quantization for generative pre-trained transformers},
  author={Frantar, Elias and Ashkboos, Saleh and Hoefler, Torsten and Alistarh, Dan},
  journal={arXiv preprint arXiv:2210.17323},
  year={2022}
}

@article{lin2024awq,
  title={Awq: Activation-aware weight quantization for on-device llm compression and acceleration},
  author={Lin, Ji and Tang, Jiaming and Tang, Haotian and Yang, Shang and Chen, Wei-Ming and Wang, Wei-Chen and Xiao, Guangxuan and Dang, Xingyu and Gan, Chuang and Han, Song},
  journal={Proceedings of machine learning and systems},
  volume={6},
  pages={87--100},
  year={2024}
}

@article{cobbe2021training,
  title={Training verifiers to solve math word problems},
  author={Cobbe, Karl and Kosaraju, Vineet and Bavarian, Mohammad and Chen, Mark and Jun, Heewoo and Kaiser, Lukasz and Plappert, Matthias and Tworek, Jerry and Hilton, Jacob and Nakano, Reiichiro and others},
  journal={arXiv preprint arXiv:2110.14168},
  year={2021}
}

@article{liu2023your,
  title={Is your code generated by chatgpt really correct? rigorous evaluation of large language models for code generation},
  author={Liu, Jiawei and Xia, Chunqiu Steven and Wang, Yuyao and Zhang, Lingming},
  journal={Advances in Neural Information Processing Systems},
  volume={36},
  pages={21558--21572},
  year={2023}
}

@article{narayan2018don,
  title={Don't give me the details, just the summary! topic-aware convolutional neural networks for extreme summarization},
  author={Narayan, Shashi and Cohen, Shay B and Lapata, Mirella},
  journal={arXiv preprint arXiv:1808.08745},
  year={2018}
}

@article{qwen2025qwen25technicalreport,
      title={Qwen2.5 Technical Report}, 
      author={An Yang and Baosong Yang and Beichen Zhang and Binyuan Hui and Bo Zheng and Bowen Yu and Chengyuan Li and Dayiheng Liu and Fei Huang and Haoran Wei and Huan Lin and Jian Yang and Jianhong Tu and Jianwei Zhang and Jianxin Yang and Jiaxi Yang and Jingren Zhou and Junyang Lin and Kai Dang and Keming Lu and Keqin Bao and Kexin Yang and Le Yu and Mei Li and Mingfeng Xue and Pei Zhang and Qin Zhu and Rui Men and Runji Lin and Tianhao Li and Tianyi Tang and Tingyu Xia and Xingzhang Ren and Xuancheng Ren and Yang Fan and Yang Su and Yichang Zhang and Yu Wan and Yuqiong Liu and Zeyu Cui and Zhenru Zhang and Zihan Qiu},
      year={2025},
      journal={arXiv preprint arXiv:2412.15115},
}

@article{jiang2023mistral7b,
      title={Mistral 7B}, 
      author={Albert Q. Jiang and Alexandre Sablayrolles and Arthur Mensch and Chris Bamford and Devendra Singh Chaplot and Diego de las Casas and Florian Bressand and Gianna Lengyel and Guillaume Lample and Lucile Saulnier and Lélio Renard Lavaud and Marie-Anne Lachaux and Pierre Stock and Teven Le Scao and Thibaut Lavril and Thomas Wang and Timothée Lacroix and William El Sayed},
      year={2023},
      journal={arXiv preprint arXiv:2310.06825},
}

@article{holtzman2019curious,
  title={The curious case of neural text degeneration},
  author={Holtzman, Ari and Buys, Jan and Du, Li and Forbes, Maxwell and Choi, Yejin},
  journal={arXiv preprint arXiv:1904.09751},
  year={2019}
}

@article{nepal2025layer,
  title={Layer Importance for Mathematical Reasoning is Forged in Pre-Training and Invariant after Post-Training},
  author={Nepal, Aadim and Shrestha, Safal and Shrestha, Anubhav and Kim, Minwu and Naghiyev, Jalal and Shwartz-Ziv, Ravid and Ross, Keith},
  journal={arXiv preprint arXiv:2506.22638},
  year={2025}
}

@article{olmo2025olmo,
  title={Olmo 3},
  author={Ettinger, Allyson and Bertsch, Amanda and Kuehl, Bailey and Graham, David and Heineman, David and Groeneveld, Dirk and Brahman, Faeze and Timbers, Finbarr and Ivison, Hamish and others},
  journal={arXiv preprint arXiv:2512.13961},
  year={2025}
}

@article{hu2024auxiliary,
  title={Auxiliary task demands mask the capabilities of smaller language models},
  author={Hu, Jennifer and Frank, Michael C},
  journal={arXiv preprint arXiv:2404.02418},
  year={2024}
}

@article{schaeffer2023emergent,
  title={Are emergent abilities of large language models a mirage?},
  author={Schaeffer, Rylan and Miranda, Brando and Koyejo, Sanmi},
  journal={Advances in neural information processing systems},
  volume={36},
  pages={55565--55581},
  year={2023}
}

@article{dettmers2023qlora,
  title={Qlora: Efficient finetuning of quantized llms},
  author={Dettmers, Tim and Pagnoni, Artidoro and Holtzman, Ari and Zettlemoyer, Luke},
  journal={Advances in neural information processing systems},
  volume={36},
  pages={10088--10115},
  year={2023}
}

@article{gao2021framework,
  title={A framework for few-shot language model evaluation},
  author={Gao, Leo and Tow, Jonathan and Biderman, Stella and Black, Sid and DiPofi, Anthony and Foster, Charles and Golding, Laurence and Hsu, Jeffrey and McDonell, Kyle and Muennighoff, Niklas and others},
  journal={Zenodo},
  year={2021}
}

@article{hendrycks2020measuring,
  title={Measuring massive multitask language understanding},
  author={Hendrycks, Dan and Burns, Collin and Basart, Steven and Zou, Andy and Mazeika, Mantas and Song, Dawn and Steinhardt, Jacob},
  journal={arXiv preprint arXiv:2009.03300},
  year={2020}
}

@inproceedings{bisk2020piqa,
  title={Piqa: Reasoning about physical commonsense in natural language},
  author={Bisk, Yonatan and Zellers, Rowan and Gao, Jianfeng and Choi, Yejin and others},
  booktitle={Proceedings of the AAAI conference on artificial intelligence},
  volume={34},
  number={05},
  pages={7432--7439},
  year={2020}
}

@article{zellers2019hellaswag,
  title={Hellaswag: Can a machine really finish your sentence?},
  author={Zellers, Rowan and Holtzman, Ari and Bisk, Yonatan and Farhadi, Ali and Choi, Yejin},
  journal={arXiv preprint arXiv:1905.07830},
  year={2019}
}

@article{sakaguchi2021winogrande,
  title={Winogrande: An adversarial winograd schema challenge at scale},
  author={Sakaguchi, Keisuke and Bras, Ronan Le and Bhagavatula, Chandra and Choi, Yejin},
  journal={Communications of the ACM},
  volume={64},
  number={9},
  pages={99--106},
  year={2021},
  publisher={ACM New York, NY, USA}
}

@article{clark2018think,
  title={Think you have solved question answering? try arc, the ai2 reasoning challenge},
  author={Clark, Peter and Cowhey, Isaac and Etzioni, Oren and Khot, Tushar and Sabharwal, Ashish and Schoenick, Carissa and Tafjord, Oyvind},
  journal={arXiv preprint arXiv:1803.05457},
  year={2018}
}

@article{mihaylov2018can,
  title={Can a suit of armor conduct electricity? a new dataset for open book question answering},
  author={Mihaylov, Todor and Clark, Peter and Khot, Tushar and Sabharwal, Ashish},
  journal={arXiv preprint arXiv:1809.02789},
  year={2018}
}

@misc{taori_alpaca_2023,
  title        = {Alpaca: {A} {Strong}, {Replicable} {Instruction}-{Following} {Model}},
  author       = {Rohan Taori and Ishaan Gulrajani and Tianyi Zhang and Yann Dubois and Xuechen Li and Carlos Guestrin and Percy Liang and Tatsunori B. Hashimoto},
  year         = {2023},
  month        = mar,
  url          = {https://crfm.stanford.edu/2023/03/13/alpaca.html},
  note         = {Stanford Center for Research on Foundation Models (CRFM)},
  urldate      = {2023-07-17}
}

@article{hinton2015distilling,
  title={Distilling the knowledge in a neural network},
  author={Hinton, Geoffrey and Vinyals, Oriol and Dean, Jeff},
  journal={arXiv preprint arXiv:1503.02531},
  year={2015}
}

\appendix

\onecolumn

\section{Appendix}

\subsection{Minitron}
\label{app:minitron_alignment}

% \subsection{Complementary Case Study on Minitron}
% \label{minitron-llama}

While our primary experiments focus on controlled recovery under resource-constrained post-training settings, we additionally analyze the \texttt{Llama-3.1-Minitron-4B-Depth} model \citep{sreenivas2024llm} as a complementary case study. This model is pruned at 50\% by selecting contiguous layers whose removal leads to minimal accuracy loss, and is subsequently trained via knowledge distillation on approximately 94B tokens using the unpruned model as the teacher. To our knowledge, it is the only publicly available checkpoint combining substantial depth pruning with large-scale post-pruning distillation.

We emphasize that Minitron is not a clean causal comparison with our controlled experiments, since it differs in pruning ratio, layer-selection procedure, training data, and post-training pipeline. We therefore treat it as suggestive external evidence. To adapt the released checkpoint to our setting, we further finetune it using our SGR-based procedure on Dolci. Despite the much larger post-pruning training budget, performance remains substantially degraded: the model achieves 23.9\% accuracy on GSM8K and 15.34\% on MBPP\textsuperscript{+}, corresponding to relative retention of 49.2\% and 34.3\%, respectively. This suggests that, for this publicly available checkpoint, large-scale distillation alone does not restore the original reasoning performance.

We note that \citet{sreenivas2024llm} report stronger performance only after additional alignment stages, including math/code supervised finetuning, instruction tuning, and preference optimization. The publicly released \texttt{Llama-3.1-Minitron-4B-Depth} checkpoint corresponds to the model after distillation (trained on $\sim$100B tokens), prior to these alignment stages. As this checkpoint is not explicitly aligned for instruction following, we perform additional finetuning using SGR before evaluating recovery, ensuring the model is adapted for reasoning-intensive tasks.

% As discussed in Appendix~\ref{app:minitron_alignment}, the released checkpoint corresponds to the post-distillation model before these stages; thus, our analysis should be interpreted as complementary evidence rather than a claim about all possible large-scale recovery pipelines.

% \citet{sreenivas2024llm} report final results after multiple post-distillation alignment stages, including supervised finetuning on math and code data, instruction tuning, and preference optimization. These stages are applied after the distillation phase.

The results in \citet{sreenivas2024llm} suggest that strong downstream performance may require not only large-scale post-pruning distillation, but also curated data and multi-stage alignment pipelines. This reinforces our focus on practical recovery settings where such large-scale data and multi-stage alignment pipelines may be unavailable.

\subsection{Tokens with Layer Pruning}
\label{tokens-text-generation}

\begin{figure}[h]
  \begin{center}
    \centerline{\includegraphics[width=0.5\columnwidth]{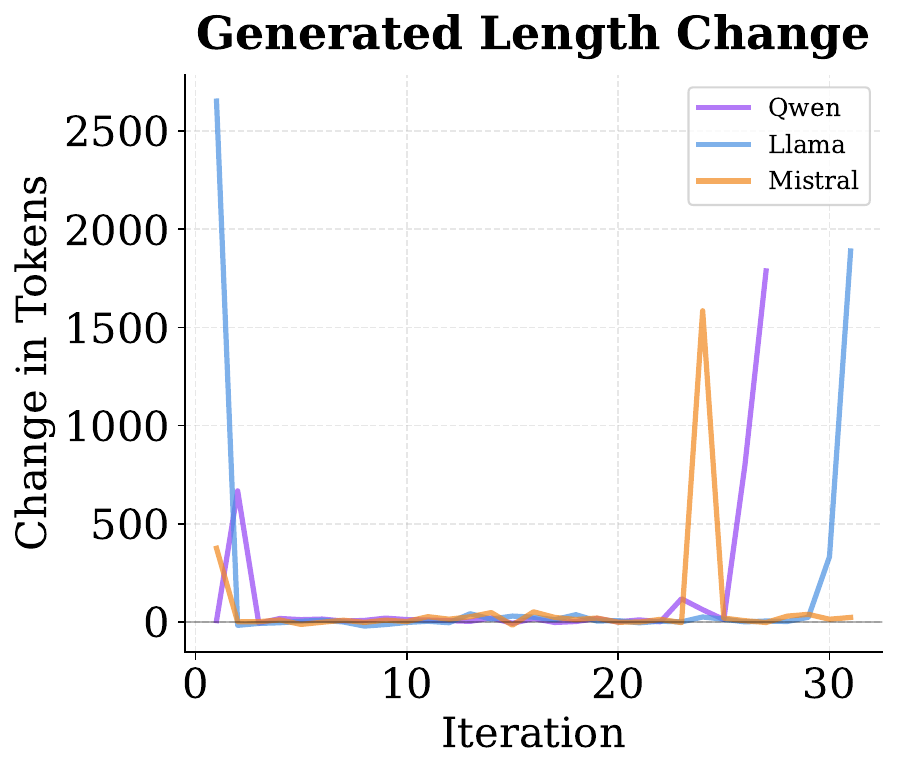}}
    \caption{Change in average generated response length after
single-layer pruning, relative to the unpruned baseline.}
    \label{fig:text-degeneration}
  \end{center}
\end{figure}

\subsection{Arithmetic Mistake}
\label{arithmetic-mistake-example}
\begin{tcolorbox}[title=Example: Weekly Egg Production]

\textbf{Question.}
Gary buys $4$ chickens. After two years, he has $8$ times as many chickens as he started with.
If each chicken lays $6$ eggs per day, how many eggs does Gary currently collect every week?

\medskip

\textbf{Pruned Model Response.}
Initially, Gary buys $4$ chickens. After two years, he has eight times as many chickens:
\[
4 \times 8 = 32 \text{ chickens.}
\]

Each chicken lays $6$ eggs per day. The total number of eggs laid per day is computed as:
\[
32 \times 6 = 364 \text{ eggs.} \qquad \textbf{(Arithmetic error)}
\]

To determine the weekly total, the daily number of eggs is multiplied by $7$:
\[
364 \times 7 = 2448 \text{ eggs.}
\]

Thus, the solution concludes that Gary collects $\boxed{2448}$ eggs per week.
\end{tcolorbox}

\subsection{Arithmetic Ablation Experiment Details}
\label{arithmetic-ablation-details}

For this experiment, we rely on the EleutherAI/arithmetic dataset. We use the single digit, three operations subset. We restrict the output space to individual digits from 0 to 9.

\begin{figure}[h!]
  \centering
  \begin{subfigure}{0.48\columnwidth}
    \centering
    \includegraphics[width=\linewidth]{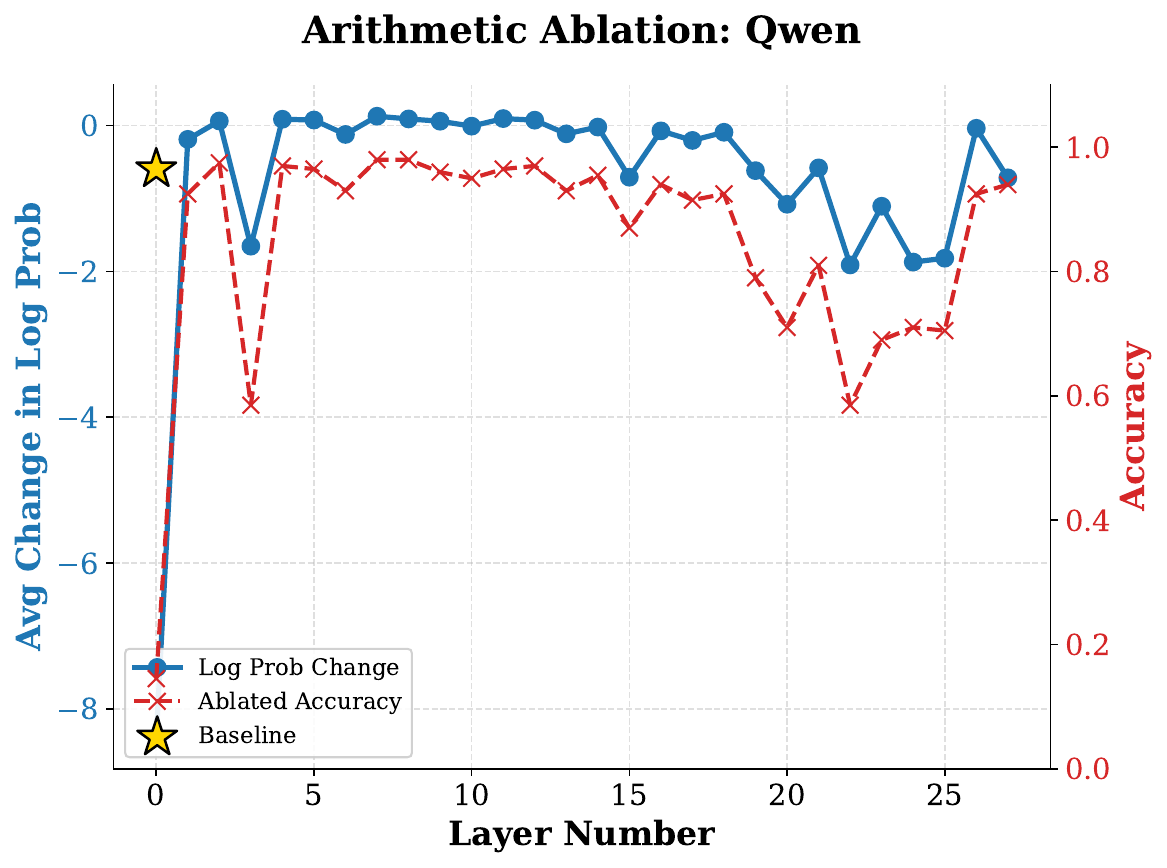}
    \caption{Qwen}
    \label{fig:qwen-arithmetic-1}
  \end{subfigure}
  \hfill
  \begin{subfigure}{0.48\columnwidth}
    \centering
    \includegraphics[width=\linewidth]{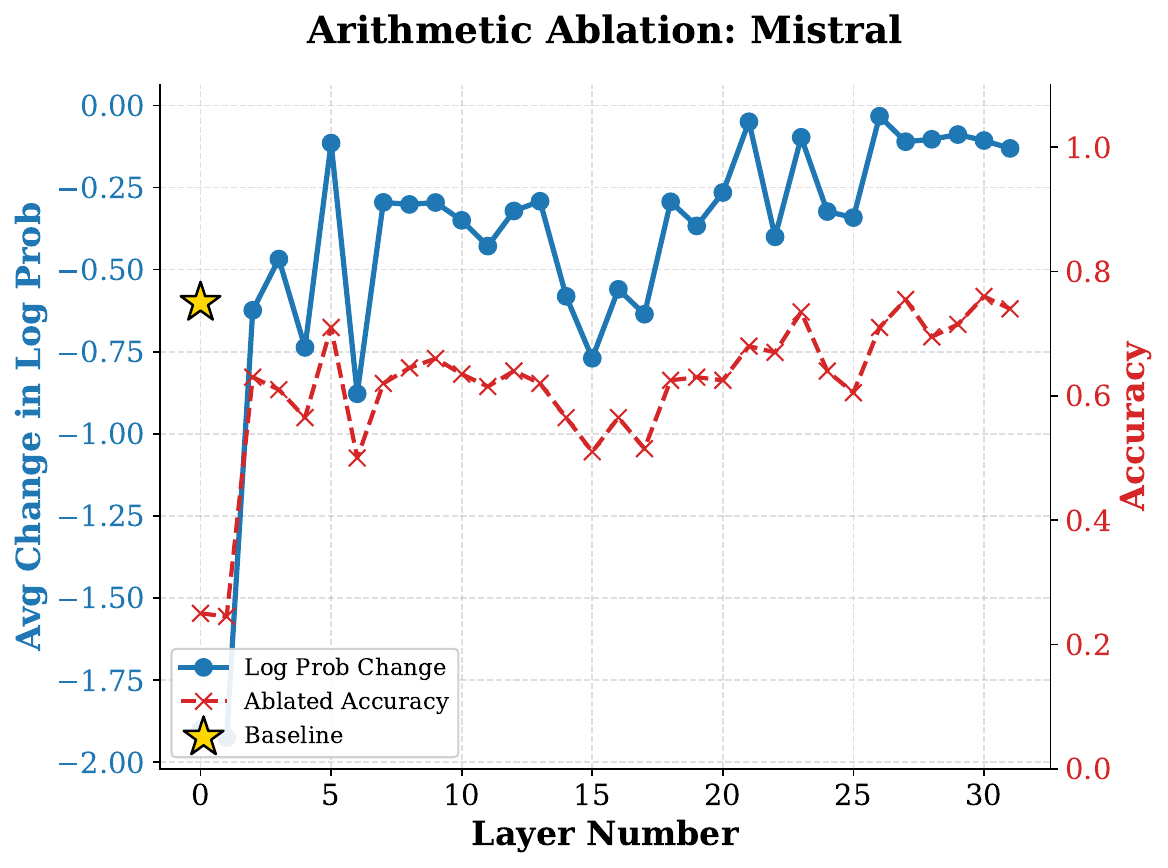}
    \caption{Mistral}
    \label{fig:mistral-arithmetic-2}
  \end{subfigure}
  \caption{
      Effect of single-layer pruning on the arithmetic ability of various models.
    }
  \label{fig:other-arithmetic-side-by-side}
\end{figure}

\subsection{Balanced Parenthesis Error}
\label{parenthesis-error}

\begin{tcolorbox}[title=Example: Code Generation with a Syntax Error]

\textbf{Prompt.}
Given an array of integers, compute the minimum number of elements that must be changed
to make the array palindromic.

\medskip

\textbf{Generated Response.}
\begin{lstlisting}[language=Python,basicstyle=\ttfamily\small]
def smallest_change(arr):
    n = len(arr)
    changes = 0
    for i in range(n // 2):
        if arr[i] != arr[n - 1 - i):
            changes += 1
        else:
            continue
    return changes
\end{lstlisting}

\medskip

\textbf{Highlighted Error.}
\[
\texttt{arr[n - 1 - i)} \qquad \textbf{(parenthesis mismatch)}
\]

\end{tcolorbox}

\subsection{Finetuning Details}
\label{all-finetuning-appendix}

\subsubsection{QLoRA with BI and Reverse Pruning Strategies}

\paragraph{Experimental Results.}
We post-train our models using QLoRA with 4-bit NF4 quantization, a learning rate of $2\times10^{-4}$, batch size 8, constant learning rate with 50 warmup steps, bf16 training, and sequence length 8192 with gradient checkpointing on a single A100 80GB GPU. We use the TRL package for training. For experiments with the Alpaca dataset ($\sim$50K), we train for 2 epochs; for Dolci \citep{olmo2025olmo} ($\sim$90K), 1 epoch. We focus on Dolci since our broader objective is to assess whether post-pruning training can preserve generative reasoning performance. We rely on QLoRA because it is comparable to even LoRA trained models in our experiments (see Table \ref{tab:pruning_external_benchmarks_all}). We also show in Appendix~\ref{qlora_vs_ft} that QLoRA closely matches performance of full finetuning for recovery in our settings as well.

\begin{table}[h]
\centering
\small
\setlength{\tabcolsep}{6pt}
\begin{tabular}{lccccccc|c}
\toprule
\textbf{Model}
& \textbf{HeSw}
& \textbf{PIQA}
& \textbf{MMLU}
& \textbf{Wino}
& \textbf{OBQA}
& \textbf{ARC-E}
& \textbf{ARC-C}
& \textbf{Mean} \\
\midrule

\multicolumn{9}{l}{\textbf{Gemma2-2B-It}} \\
\midrule
Reverse-order*
& 0.844 & 0.893 & 0.925 & 0.941 & 0.747 & 0.785 & 0.736 & \underline{0.839} \\
PPL*
& 0.859 & 0.948 & 0.616 & 0.837 & 0.826 & 0.867 & 0.706 & 0.808 \\
Magnitude-L1*
& 0.844 & 0.951 & 0.475 & 0.804 & 0.854 & 0.895 & 0.730 & 0.793 \\
Magnitude-L2*
& 0.791 & 0.918 & 0.424 & 0.789 & 0.669 & 0.812 & 0.594 & 0.714 \\
BI*
& 0.796 & 0.880 & 0.874 & 0.957 & 0.758 & 0.805 & 0.725 & 0.828 \\
Taylor*
& 0.846 & 0.890 & 0.955 & 0.932 & 0.848 & 0.787 & 0.723 & \textbf{0.854} \\[1.5mm]

\textit{QLoRA (Reverse + Dolci)}
& \textit{0.736} & \textit{0.857} & \textit{0.930} & \textit{0.945} & \textit{0.702} & \textit{0.816} & \textit{0.708} & \textit{0.813} \\
\textit{QLoRA (BI + Dolci)} &
\textit{0.729} & \textit{0.857} & \textit{0.856} & \textit{0.961} & \textit{0.927} & \textit{0.774} & \textit{0.698} & \textit{0.83} \\

\midrule
\multicolumn{9}{l}{\textbf{LLaMA-3.1-8B-It}} \\
\midrule
Reverse-order*
& 0.679 & 0.875 & 0.934 & 0.844 & 0.870 & 0.754 & 0.769 & 0.818 \\
PPL*
& 0.834 & 0.953 & 0.496 & 0.783 & 0.781 & 0.891 & 0.735 & 0.782 \\
Magnitude-L1*
& 0.446 & 0.676 & 0.369 & 0.660 & 0.402 & 0.348 & 0.389 & 0.470 \\
Magnitude-L2*
& 0.446 & 0.676 & 0.368 & 0.659 & 0.396 & 0.347 & 0.389 & 0.469 \\
BI*
& 0.710 & 0.897 & 0.356 & 0.729 & 0.598 & 0.746 & 0.549 & 0.655 \\
Taylor*
& 0.840 & 0.892 & 0.421 & 0.965 & 0.811 & 0.837 & 0.807 & 0.796 \\[1.5mm]

\textit{QLoRA (Reverse + Dolci)}
& \textit{0.787} & \textit{0.896} & \textit{0.895} & \textit{1.000} & \textit{0.651} & \textit{0.863} & \textit{0.813} & \textbf{\textit{0.844}} \\
\textit{QLoRA (BI + Dolci)}
& \textit{0.689} & \textit{0.862} & \textit{0.925} & \textit{0.915} & \textit{0.858} & \textit{0.786} & \textit{0.780} & \textit{\underline{0.831}} \\
\bottomrule
\end{tabular}
\caption{
Performance retention on classification benchmarks across Gemma and Llama with various pruning strategies. LoRA-trained results from \citep{lu2024reassessing} are marked with an asterisk (*). QLoRA results largely match or outperform LoRA trained models across various pruning strategies showing consistent $>$80\% performance retention. HeSw = HellaSwag, Wino = Winogrande, OBQA = OpenBookQA.
}
\label{tab:pruning_external_benchmarks_all}
\end{table}

\paragraph{Pruning Strategies.}
In our experiments, we mainly deal with two commonly used layer pruning strategies, Block Influence (BI) and Reverse Order \citep{men2025shortgpt,lu2024reassessing,sreenivas2024llm}. We show that QLoRA with simple pruning metrics like BI and Reverse perform comparably with techniques like LoRA composed with other various pruning metrics (see Table \ref{tab:pruning_external_benchmarks_all}). Additionally, we consider an \emph{iterative} pruning procedure. The iterative procedure extends the single-layer pruning analysis in Figure~\ref{fig:multi-model-ablation} by greedily removing layers based on observed redundancy. Specifically, we first prune the layer whose removal leads to the smallest performance drop. With this layer removed, we then determine the layer whose removal leads to the smallest drop, and additionally prune that layer. We continue to repeat this process. This pruning strategy allows us to examine whether selectively removing redundant layers affects the recovery behavior observed in generative reasoning tasks. The full procedure is described in Algorithm~\ref{alg:greedy_pruning}.

\begin{algorithm}[h]
\caption{Greedy Iterative Pruning via Benchmark Performance}
\label{alg:greedy_pruning}
\begin{algorithmic}
\STATE \textbf{Input:} Model $\mathcal{M}$ with $L$ layers, benchmark dataset $\mathcal{D}$, number of layers to prune $N$
\STATE \textbf{Output:} Pruned layer set $\mathcal{P}$

\STATE $\mathcal{P} \leftarrow \emptyset$
\FOR{$k = 1$ to $N$}
    \STATE $\ell^\star \leftarrow 
    \arg\max_{\ell \in \{1,\dots,L\} \setminus \mathcal{P}}
    \; \text{Score}\!\left(\mathcal{M}_{-(\mathcal{P} \cup \{\ell\})}, \mathcal{D}\right)$
    \STATE $\mathcal{P} \leftarrow \mathcal{P} \cup \{\ell^\star\}$
\ENDFOR
\STATE \textbf{Return} $\mathcal{P}$
\end{algorithmic}
\end{algorithm}

\subsection{Task-Aligned Recovery on GSM8K}
\label{upper-bound}

\begin{figure}[h]
  \begin{center}
    \centerline{\includegraphics[width=0.5\columnwidth]{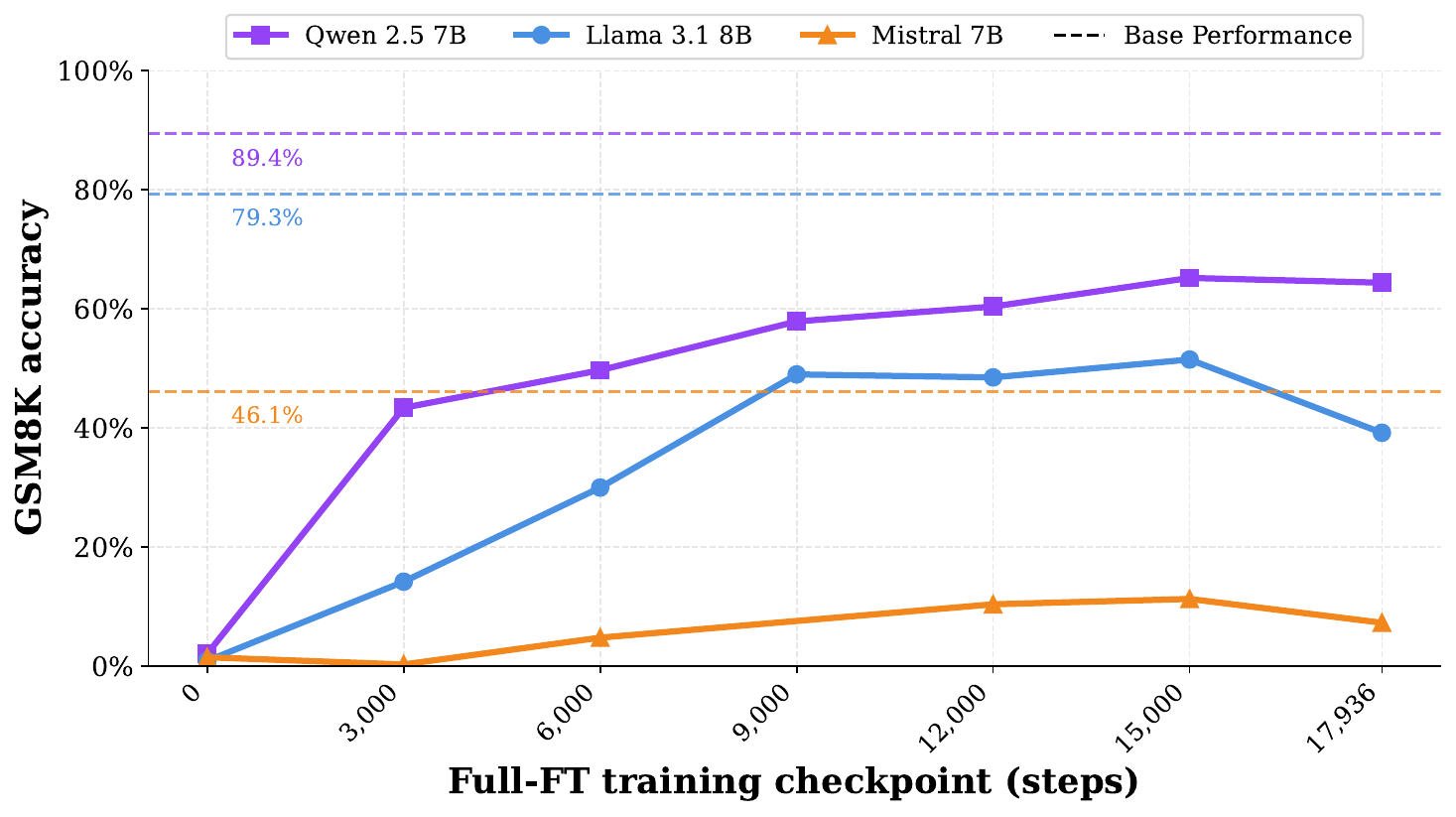}}
    \caption{
      Full Finetuning with GSM8K dataset.
    }
    \label{fig:qlora_vs_sft}
  \end{center}
\end{figure}

To test whether the recovery gap persists under a favorable task-aligned recovery setting, we conduct full-parameter finetuning experiments on GSM8K after pruning. We remove 7 layers from Qwen and 8 layers each from LLaMA and Mistral. For all models, we generate self-generated responses from the corresponding unpruned model on the GSM8K training split, finetune the pruned model on this task-specific data, and evaluate on the GSM8K test set. This setting is more favorable than our main recovery experiments because the recovery data and evaluation task are aligned, and recovery is performed with full finetuning rather than QLoRA. For Qwen, we use the Iterative pruning procedure with GSM8K train data as the calibration benchmark, making the pruning selection itself task-aligned. For LLaMA and Mistral, we use the cosine-similarity pruning metric. We include both settings to test whether task-specific full finetuning can restore reasoning performance after depth reduction across model families.

\paragraph{Training details.}
For this task-aligned recovery experiment, we perform full-parameter
supervised finetuning of the layer-pruned Llama-3.1-8B-Instruct,
Mistral-7B-Instruct-v0.3, and Qwen2.5-7B-Instruct models. For each
model, its corresponding unpruned checkpoint generates eight responses
for every problem in the GSM8K training split, yielding 59,784
prompt--response examples. We randomly hold out 10\% of these examples
for validation, resulting in 53,806 training examples and 5,978
validation examples. The training loss is computed only over
response-completion tokens.

We train each pruned model for two epochs (17,936 optimization steps)
using three A100 80GB GPUs with ZeRO-3 distributed training. Training
uses bfloat16 precision, a per-device batch size of 2, gradient
accumulation of 1, an effective global batch size of 6, a maximum
sequence length of 8,192 tokens, and gradient checkpointing. We
optimize with AdamW using a learning rate of $2\times10^{-5}$,
$\beta_1=0.9$, $\beta_2=0.95$, weight decay of $10^{-4}$, gradient
clipping at 1.0, a 3\% warmup ratio, and cosine learning-rate decay.
We use random seed 42 and save checkpoints every 3,000 optimization
steps. Unlike our main resource-constrained SGR experiments, which can
be conducted on a single 80GB GPU, this task-aligned full-finetuning
experiment uses three GPUs as a stronger recovery check.

\subsubsection{QLoRA vs.\ Full Finetuning}
\label{qlora_vs_ft}

\begin{figure}[h]
  \begin{center}
    \centerline{\includegraphics[width=0.5\columnwidth]{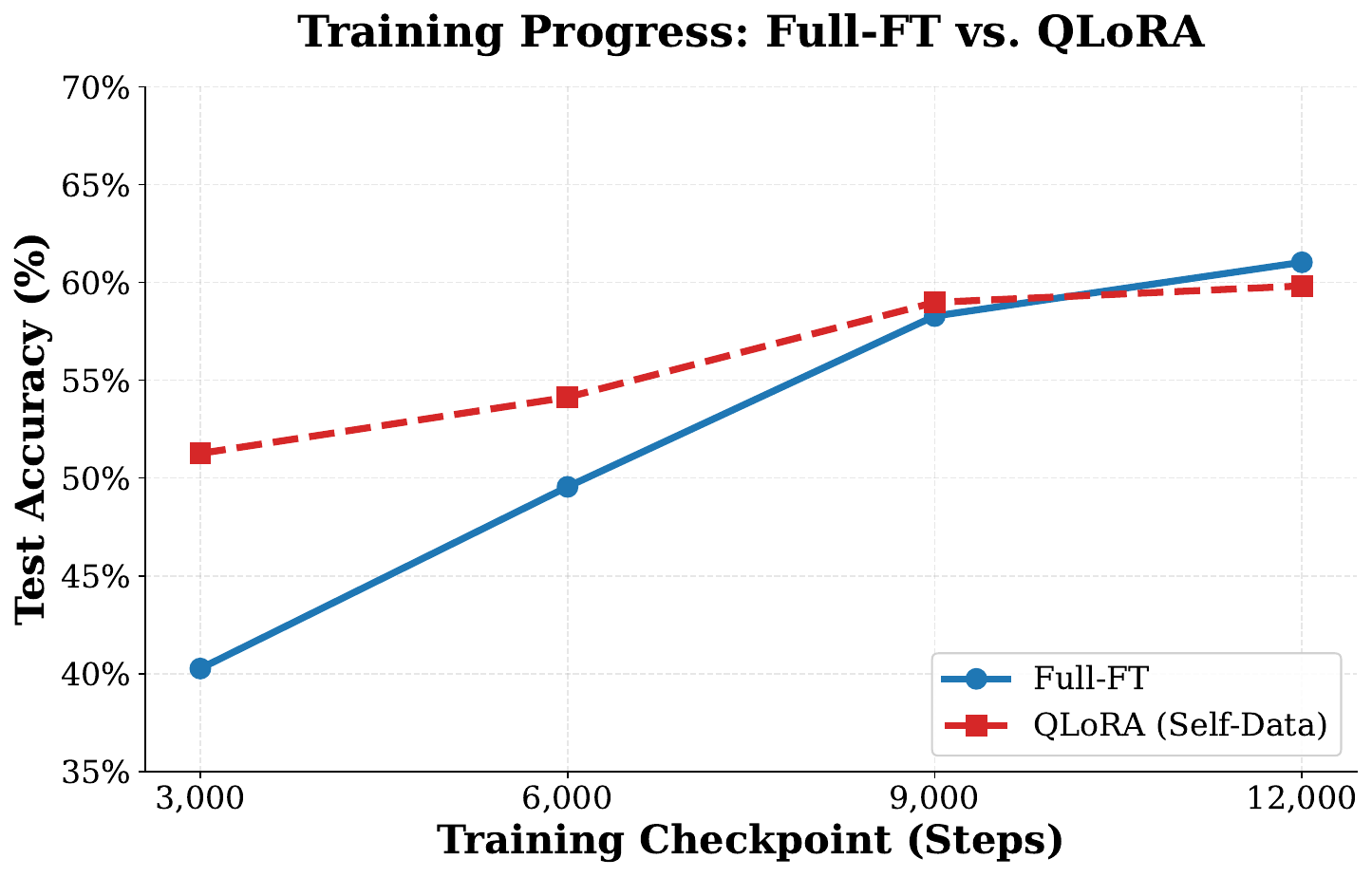}}
    \caption{
      Comparison between full supervised finetuning (Full-FT) and QLoRA on self-generated Dolci data.
    }
    \label{fig:ft_vs_qlora}
  \end{center}
\end{figure}

We further compare QLoRA against full-parameter supervised finetuning (Full-FT) using self-generated Dolci data. Both methods are applied to the same Iteratively pruned Qwen model with seven layers removed, and evaluated on GSM8K as a proxy for generative reasoning quality.

At the scale of our experiments, QLoRA achieves recovery comparable to Full-FT, despite operating under significantly reduced memory and compute requirements. While full finetuning may yield additional gains when scaled further, doing so incurs substantially higher computational cost and hardware demands. Given our focus on recovery under resource-constrained post-training settings, we adopt QLoRA throughout the paper as a practical and representative finetuning approach for studying the limits of generative reasoning recovery after layer pruning.

\subsection{RLVR-Based Recovery on GSM8K}
\label{rlvr-recovery-appendix}
To assess whether reinforcement-based alignment can further recover
capabilities degraded by depth reduction, we apply RLVR finetuning on
GSM8K to the SGR-recovered Qwen2.5-7B-Instruct and
Mistral-7B-Instruct-v0.3 models, each pruned by seven layers using Block
Influence (BI).

As shown in Table~\ref{tab:rlvr-recovery}, RLVR improves Qwen's GSM8K
retention from 0.329 after SGR to 0.418. For Mistral, the best RLVR
checkpoint achieves 28.51\% GSM8K accuracy. Performance remains near
this level through 1,500 training steps before declining sharply with
continued optimization, suggesting sensitivity to training duration.
Together, these results indicate that targeted reinforcement learning
can partially improve post-pruning mathematical reasoning, although the
gains do not close the gap with the unpruned models and may be unstable
during extended training.

\begin{table}[h!]
  \centering
  \small
  \begin{tabular}{llc}
    \toprule
    \textbf{Model} & \textbf{Recovery Strategy} & \textbf{GSM8K Retention} \\
    \midrule
    \multirow{4}{*}{Qwen2.5-7B-Instruct}
      & Unpruned baseline & 1.000 \\
      & BI pruning + SFT on Dolci & 0.270 \\
      & BI pruning + SGR on S.Dolci & 0.329 \\
      & BI pruning + SGR + RLVR & 0.418 \\
    \midrule
    \multirow{3}{*}{Mistral-7B-Instruct-v0.3}
      & BI pruning + SFT on Dolci & 0.236 \\
      & BI pruning + SGR on S.Dolci & 0.395 \\
      & BI pruning + SGR + RLVR & 0.61 \\
    \bottomrule
  \end{tabular}
  \caption{GSM8K results under successive post-pruning RLVR for Qwen2.5-7B-Instruct and Mistral-7B-Instruct-v0.3,
  each with seven layers removed using BI. Qwen and the Mistral results are reported as retention normalized to their
  respective unpruned baselines.}
  \label{tab:rlvr-recovery}
\end{table}

\subsection{Perplexity Curves}
\label{ppl-curves-appendix}

\begin{figure}[h]
  \centering
  \begin{subfigure}{0.48\columnwidth}
    \centering
    \includegraphics[width=\linewidth]{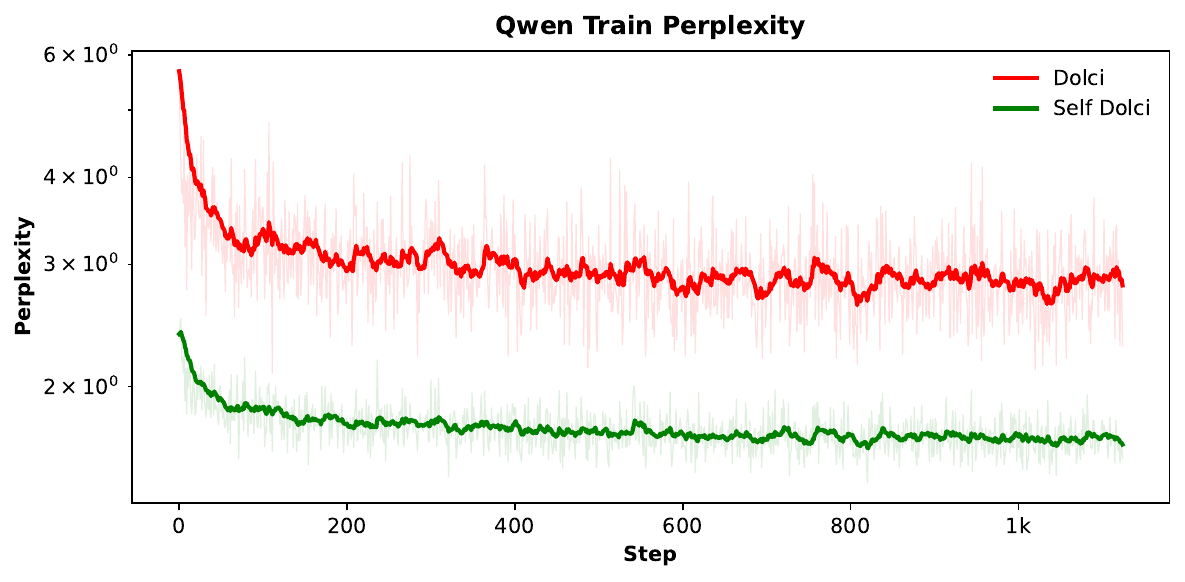}
    \caption{Qwen}
    \label{fig:qwen-ppl}
  \end{subfigure}
  \hfill
  \begin{subfigure}{0.48\columnwidth}
    \centering
    \includegraphics[width=\linewidth]{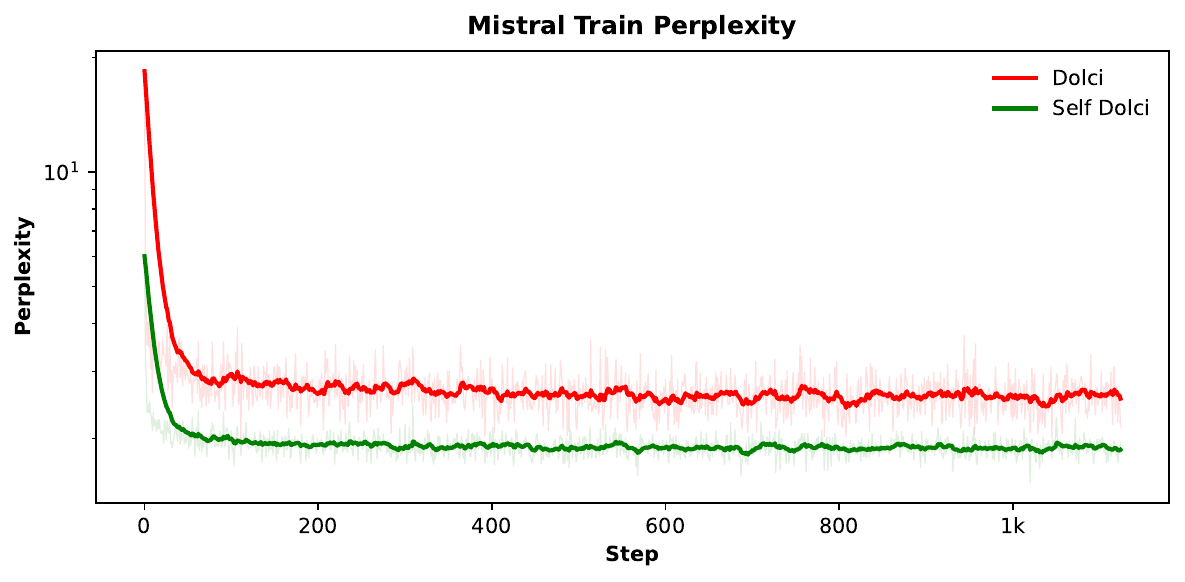}
    \caption{Mistral}
    \label{fig:mistral-ppl}
  \end{subfigure}
  \caption{
      Perplexity curves during training for both standard finetuning and for SGR for the Qwen and Mistral models (Both are BI pruned: 25\%).
    }
  \label{fig:appendix-ppl}
\end{figure}

\subsection{Full Results with Self-Generated Responses}
\label{sec:self_generated_full}
 The full results are in Table \ref{tab:retention_joint_full}. We also observe that, in Table \ref{tab:retention_joint_full}, the effectiveness of pruning strategies varies by model. While Iterative pruning does not provide consistent gains over standard metrics such as BI and Reverse Order for Llama and Mistral, it performs better for Qwen. This suggests that greedily selecting layers based on empirical redundancy does not generalize uniformly across architectures, potentially reflecting differences in internal layer organization, as previously noted for Qwen models \citep{gromov2024unreasonable}.

\begin{table}[h]
\centering
\scriptsize
\setlength{\tabcolsep}{3.5pt}
\renewcommand{\arraystretch}{0.95}
\resizebox{\textwidth}{!}{
\begin{tabular}{l ccccccc c @{\hspace{3pt}}|@{\hspace{3pt}} cccc c}
\toprule
& \multicolumn{8}{c}{\textbf{Classification}}
& \multicolumn{5}{c}{\textbf{Generative}} \\
\cmidrule(lr){2-9}\cmidrule(lr){10-14}
\textbf{Model}
& \textbf{HeSw}
& \textbf{PIQA}
& \textbf{MMLU}
& \textbf{Wino}
& \textbf{OBQA}
& \textbf{ARC-E}
& \textbf{ARC-C}
& \textbf{Mean}
& \textbf{GSM8K}
& \textbf{HumEval$^+$}
& \textbf{MBPP$^+$}
& \textbf{XSUM}
& \textbf{Mean} \\
\midrule

\multicolumn{14}{l}{\textbf{Gemma2-2B-It}} \\
\midrule
\multicolumn{14}{l}{\textbf{\emph{Open-source data}}} \\
Reverse + Alpaca 
& 0.844 & 0.893 & 0.925 & 0.941 & 0.747 & 0.785 & 0.736 & \underline{0.839}
& 0.040 & 0.119 & 0.029 & 0.814 & 0.251 \\

BI + Alpaca
& 0.796 & 0.880 & 0.874 & 0.957 & 0.758 & 0.805 & 0.725 & 0.828
& 0.042 & 0.153 & 0.099 & 0.809 & 0.276 \\

Reverse + Dolci
& 0.736 & 0.857 & 0.930 & 0.945 & 0.702 & 0.816 & 0.708 & 0.813
& 0.129 & 0.219 & 0.121 & 0.763 & \underline{0.308} \\

BI + Dolci
& 0.729 & 0.857 & 0.856 & 0.961 & 0.927 & 0.774 & 0.698 & 0.829
& 0.154 & 0.095 & 0.113 & 0.778 & 0.285 \\

\addlinespace[0.5mm]
\multicolumn{14}{l}{\textbf{\emph{Self-Generated Responses}}} \\
\textit{SGR (Reverse + S.Alpaca)}
& 0.783 & 0.861 & 0.930 & 0.984 & 0.702 & 0.866 & 0.849 & 0.854~\textcolor{darkgreen}{\boldmath$\uparrow$}
& 0.047 & 0.103 & 0.212 & 0.856 & 0.304~\textcolor{darkgreen}{\boldmath$\uparrow$} \\

\textit{SGR (BI + S.Alpaca)}
& 0.756 & 0.847 & 0.874 & 0.888 & 0.843 & 0.866 & 0.849 & 0.846~\textcolor{darkgreen}{\boldmath$\uparrow$}
& 0.067 & 0.186 & 0.164 & 0.851 & 0.317~\textcolor{darkgreen}{\boldmath$\uparrow$} \\

\textit{SGR (Reverse + S.Dolci)}
& 0.732 & 0.805 & 0.937 & 0.940 & 0.758 & 0.845 & 0.824 & 0.834~\textcolor{darkgreen}{\boldmath$\uparrow$}
& 0.164 & 0.256 & 0.121 & 0.768 & 0.327~\textcolor{darkgreen}{\boldmath$\uparrow$} \\

\textit{SGR (BI + S.Dolci)}
& 0.729 & 0.857 & 0.904 & 0.945 & 0.955 & 0.821 & 0.800 & \textbf{0.859~\textcolor{darkgreen}{\boldmath$\uparrow$}}
& 0.259 & 0.389 & 0.298 & 0.876 & \textbf{0.455~\textcolor{darkgreen}{\boldmath$\uparrow$}} \\

\midrule
\multicolumn{14}{l}{\textbf{LLaMA-3.1-8B-It}} \\
\midrule
\multicolumn{14}{l}{\textbf{\emph{Open-source data}}} \\
Reverse + Alpaca*
& 0.679 & 0.875 & 0.934 & 0.844 & 0.870 & 0.754 & 0.769 & 0.818
& 0.453 & 0.111 & 0.084 & 0.638 & 0.321 \\

BI + Alpaca
& 0.710 & 0.897 & 0.356 & 0.729 & 0.598 & 0.746 & 0.549 & 0.655
& 0.318 & 0.344 & 0.128 & 0.062 & 0.213 \\

Reverse + Dolci
& 0.787 & 0.896 & 0.895 & 1.000 & 0.651 & 0.863 & 0.813 & 0.844
& 0.405 & 0.434 & 0.302 & 0.077 & 0.304 \\

BI + Dolci
& 0.793 & 0.886 & 0.943 & 1.027 & 0.710 & 0.896 & 0.887 & \underline{0.878}
& 0.469 & 0.444 & 0.308 & 0.068 & \underline{0.322} \\

Iterative + Dolci
& 0.787 & 0.896 & 0.789 & 0.926 & 0.828 & 0.893 & 0.747 & 0.838
& 0.328 & 0.301 & 0.245 & 0.338 & 0.303 \\

\addlinespace[0.5mm]
\multicolumn{14}{l}{\textbf{\emph{Self-Generated Responses}}} \\
\textit{SGR (Reverse + S.Alpaca)}
& 0.682 & 0.860 & 0.909 & 0.926 & 0.947 & 0.807 & 0.772 & 0.843~\textcolor{darkgreen}{\boldmath$\uparrow$}
& 0.561 & 0.290 & 0.162 & 0.179 & 0.298~\textcolor{darkred}{\boldmath$\downarrow$} \\

\textit{SGR (BI + S.Alpaca)}
& 0.838 & 0.913 & 0.967 & 1.049 & 0.769 & 0.950 & 0.912 & \textbf{0.914~\textcolor{darkgreen}{\boldmath$\uparrow$}}
& 0.724 & 0.412 & 0.481 & 0.860 & 0.619~\textcolor{darkgreen}{\boldmath$\uparrow$} \\

\textit{SGR (Reverse + S.Dolci)}
& 0.809 & 0.920 & 0.963 & 0.979 & 0.799 & 0.917 & 0.879 & 0.895~\textcolor{darkgreen}{\boldmath$\uparrow$}
& 0.628 & 0.556 & 0.251 & 0.029 & 0.366~\textcolor{darkgreen}{\boldmath$\uparrow$} \\

\textit{SGR (BI + S.Dolci)}
& 0.826 & 0.907 & 0.984 & 1.033 & 0.769 & 0.924 & 0.879 & 0.903~\textcolor{darkgreen}{\boldmath$\uparrow$}
& 0.758 & 0.634 & 0.390 & 0.754 & \textbf{0.634~\textcolor{darkgreen}{\boldmath$\uparrow$}} \\

\textit{SGR (Iterative + S.Dolci)}
& 0.792 & 0.907 & 0.835 & 1.022 & 0.947 & 0.875 & 0.780 & 0.880~\textcolor{darkgreen}{\boldmath$\uparrow$}
& 0.647 & 0.423 & 0.312 & 0.763 & 0.536~\textcolor{darkgreen}{\boldmath$\uparrow$} \\

\midrule
\multicolumn{14}{l}{\textbf{Qwen2.5-7B-Instruct}} \\
\midrule
\multicolumn{14}{l}{\textbf{\emph{Open-source data}}} \\
Reverse + Alpaca
& 0.710 & 0.832 & 0.765 & 0.854 & 0.488 & 0.720 & 0.570 & 0.706
& 0.012 & 0.025 & 0.020 & 0.591 & 0.162 \\

BI + Alpaca
& 0.867 & 0.982 & 0.480 & 0.789 & 0.927 & 0.874 & 0.708 & \underline{0.804}
& 0.097 & 0.167 & 0.274 & 0.768 & 0.327 \\

Reverse + Dolci
& 0.681 & 0.794 & 0.703 & 0.826 & 0.488 & 0.650 & 0.524 & 0.666
& 0.059 & 0.159 & 0.120 & 0.541 & 0.220 \\

BI + Dolci
& 0.820 & 0.958 & 0.507 & 0.854 & 0.878 & 0.843 & 0.626 & 0.784
& 0.270 & 0.200 & 0.278 & 0.817 & 0.391 \\

Iterative + Dolci
& 0.827 & 0.910 & 0.700 & 0.859 & 0.732 & 0.828 & 0.727 & 0.798
& 0.294 & 0.333 & 0.358 & 0.812 & \underline{0.449} \\

\addlinespace[0.5mm]
\multicolumn{14}{l}{\textbf{\emph{Self-Generated Responses}}} \\
\textit{SGR (Reverse + S.Alpaca)}
& 0.740 & 0.856 & 0.848 & 0.846 & 0.488 & 0.741 & 0.633 & 0.736~\textcolor{darkgreen}{\boldmath$\uparrow$}
& 0.025 & 0.051 & 0.024 & 0.605 & 0.176~\textcolor{darkgreen}{\boldmath$\uparrow$} \\

\textit{SGR (BI + S.Alpaca)}
& 0.891 & 0.982 & 0.540 & 0.826 & 0.732 & 0.852 & 0.748 & 0.796~\textcolor{darkred}{\boldmath$\downarrow$}
& 0.202 & 0.183 & 0.288 & 0.846 & 0.380~\textcolor{darkgreen}{\boldmath$\uparrow$} \\

\textit{SGR (Reverse + S.Dolci)}
& 0.678 & 0.788 & 0.731 & 0.732 & 0.516 & 0.652 & 0.576 & 0.668~\textcolor{darkgreen}{\boldmath$\uparrow$}
& 0.153 & 0.142 & 0.157 & 0.600 & 0.263~\textcolor{darkgreen}{\boldmath$\uparrow$} \\

\textit{SGR (BI + S.Dolci)}
& 0.851 & 0.982 & 0.545 & 0.841 & 0.854 & 0.867 & 0.693 & \textbf{0.805~\textcolor{darkgreen}{\boldmath$\uparrow$}}
& 0.329 & 0.283 & 0.398 & 0.861 & 0.468~\textcolor{darkgreen}{\boldmath$\uparrow$} \\

\textit{SGR (Iterative + S.Dolci)}
& 0.852 & 0.934 & 0.699 & 0.852 & 0.610 & 0.869 & 0.775 & 0.799~\textcolor{darkgreen}{\boldmath$\uparrow$}
& 0.581 & 0.325 & 0.374 & 0.822 & \textbf{0.525}~\textcolor{darkgreen}{\boldmath$\uparrow$} \\

\midrule
\multicolumn{14}{l}{\textbf{Mistralv0.3-7B-Instruct}} \\
\midrule
\multicolumn{14}{l}{\textbf{\emph{Open-source data}}} \\
Reverse + Alpaca
& 0.832 & 0.863 & 0.769 & 0.886 & 0.523 & 0.849 & 0.799 & 0.789
& 0.087 & 0.158 & 0.246 & 0.096 & 0.147 \\

BI + Alpaca
& 0.827 & 0.848 & 0.823 & 0.896 & 0.705 & 0.843 & 0.663 & \underline{0.801}
& 0.056 & 0.244 & 0.223 & 0.048 & 0.143 \\

Reverse + Dolci
& 0.792 & 0.834 & 0.844 & 0.941 & 0.591 & 0.821 & 0.601 & 0.775
& 0.375 & 0.316 & 0.488 & 0.108 & \underline{0.322} \\

BI + Dolci
& 0.813 & 0.848 & 0.877 & 0.951 & 0.614 & 0.786 & 0.608 & 0.785
& 0.236 & 0.316 & 0.285 & 0.072 & 0.227 \\

Iterative + Dolci
& 0.806 & 0.886 & 0.685 & 0.851 & 0.682 & 0.804 & 0.621 & 0.762
& 0.293 & 0.244 & 0.262 & 0.096 & 0.224 \\

\addlinespace[0.5mm]
\multicolumn{14}{l}{\textbf{\emph{Self-Generated Responses}}} \\
\textit{SGR (Reverse + S.Alpaca)}
& 0.876 & 0.890 & 0.891 & 0.866 & 0.773 & 0.903 & 0.812 & \textbf{0.859~\textcolor{darkgreen}{\boldmath$\uparrow$}}
& 0.182 & 0.282 & 0.569 & 0.054 & 0.272~\textcolor{darkgreen}{\boldmath$\uparrow$} \\

\textit{SGR (BI + S.Alpaca)}
& 0.860 & 0.873 & 0.924 & 0.856 & 0.614 & 0.851 & 0.751 & 0.818~\textcolor{darkgreen}{\boldmath$\uparrow$}
& 0.102 & 0.264 & 0.569 & 0.084 & 0.255~\textcolor{darkgreen}{\boldmath$\uparrow$} \\

\textit{SGR (Reverse + S.Dolci)}
& 0.861 & 0.863 & 0.929 & 0.906 & 0.750 & 0.883 & 0.758 & 0.850~\textcolor{darkgreen}{\boldmath$\uparrow$}
& 0.421 & 0.526 & 0.875 & 0.095 & \textbf{0.479~\textcolor{darkgreen}{\boldmath$\uparrow$}} \\

\textit{SGR (BI + S.Dolci)}
& 0.854 & 0.857 & 0.908 & 0.911 & 0.614 & 0.831 & 0.772 & 0.821~\textcolor{darkgreen}{\boldmath$\uparrow$}
& 0.395 & 0.509 & 0.715 & 0.066 & 0.421~\textcolor{darkgreen}{\boldmath$\uparrow$} \\

\textit{SGR (Iterative + S.Dolci)}
& 0.841 & 0.871 & 0.718 & 0.836 & 0.682 & 0.856 & 0.657 & 0.780~\textcolor{darkgreen}{\boldmath$\uparrow$}
& 0.230 & 0.402 & 0.492 & 0.042 & 0.292~\textcolor{darkgreen}{\boldmath$\uparrow$} \\

\bottomrule
\end{tabular}
}
\caption{
Performance retention (normalized to baseline). Results marked with (*) are sourced from \citep{lu2024reassessing}.
SGR is our method with the pruning metric in parentheses. Reverse, BI, and Iterative (described in Algorithm \ref{alg:greedy_pruning}) indicate the pruning order, while Alpaca and Dolci denote the training data source; S.Alpaca and S.Dolci refer to self-generated variants of the corresponding datasets.
$\uparrow/\downarrow$ indicate improvement or degradation in mean retention of our SGR approach with respect to the standard approach of doing SFT with the open-source prompts and responses.
}
\label{tab:retention_joint_full}
\end{table}

\begin{table}[h]
\centering
\scriptsize
\setlength{\tabcolsep}{3.5pt}
\renewcommand{\arraystretch}{0.95}
\resizebox{\textwidth}{!}{
\begin{tabular}{l ccccccc c @{\hspace{3pt}}|@{\hspace{3pt}} cccc c}
\toprule
& \multicolumn{8}{c}{\textbf{Classification}}
& \multicolumn{5}{c}{\textbf{Generative}} \\
\cmidrule(lr){2-9}\cmidrule(lr){10-14}
\textbf{Model}
& \textbf{HeSw}
& \textbf{PIQA}
& \textbf{MMLU}
& \textbf{Wino}
& \textbf{OBQA}
& \textbf{ARC-E}
& \textbf{ARC-C}
& \textbf{Mean}
& \textbf{GSM8K}
& \textbf{HumEval$^+$}
& \textbf{MBPP$^+$}
& \textbf{XSUM}
& \textbf{Mean} \\
\midrule

\multicolumn{14}{l}{\textbf{Gemma2-2B-It}} \\
\midrule
\multicolumn{14}{l}{\textbf{\emph{Open-source data}}} \\

Reverse + Alpaca
& 0.453 & 0.703 & 0.526 & 0.655 & 0.266 & 0.635 & 0.376 & 0.516
& 0.024 & 0.043 & 0.011 & 0.158 & 0.059 \\

BI + Alpaca
& 0.427 & 0.692 & 0.497 & 0.666 & 0.270 & 0.651 & 0.370 & 0.511
& 0.025 & 0.055 & 0.037 & 0.157 & 0.069 \\

Reverse + Dolci
& 0.395 & 0.674 & 0.529 & 0.658 & 0.250 & 0.660 & 0.362 & 0.504
& 0.077 & 0.079 & 0.045 & 0.148 & 0.087 \\

BI + Dolci
& 0.391 & 0.674 & 0.487 & 0.669 & 0.330 & 0.626 & 0.357 & 0.505
& 0.092 & 0.034 & 0.042 & 0.151 & 0.080 \\

\addlinespace[0.5mm]
\multicolumn{14}{l}{\textbf{\emph{Self-Generated Responses}}} \\

\textit{SGR (Reverse + S.Alpaca)}
& 0.420 & 0.677 & 0.529 & 0.685 & 0.250 & 0.700 & 0.434 & 0.528
& 0.028 & 0.037 & 0.079 & 0.166 & 0.078 \\

\textit{SGR (BI + S.Alpaca)}
& 0.406 & 0.666 & 0.497 & 0.618 & 0.300 & 0.700 & 0.434 & 0.517
& 0.040 & 0.067 & 0.061 & 0.165 & 0.083 \\

\textit{SGR (Reverse + S.Dolci)}
& 0.393 & 0.633 & 0.533 & 0.654 & 0.270 & 0.683 & 0.421 & 0.512
& 0.098 & 0.092 & 0.045 & 0.149 & 0.096 \\

\textit{SGR (BI + S.Dolci)}
& 0.391 & 0.674 & 0.514 & 0.658 & 0.340 & 0.664 & 0.409 & 0.521
& 0.155 & 0.140 & 0.111 & 0.170 & 0.144 \\

\midrule
\multicolumn{14}{l}{\textbf{LLaMA-3.1-8B-It}} \\
\midrule
\multicolumn{14}{l}{\textbf{\emph{Open-source data}}} \\

Reverse + Alpaca*
& 0.401 & 0.700 & 0.634 & 0.624 & 0.294 & 0.617 & 0.398 & 0.524
& 0.359 & 0.061 & 0.040 & 0.132 & 0.148 \\

BI + Alpaca
& 0.420 & 0.718 & 0.242 & 0.539 & 0.202 & 0.610 & 0.284 & 0.431
& 0.252 & 0.189 & 0.061 & 0.013 & 0.129 \\

Reverse + Dolci
& 0.465 & 0.717 & 0.608 & 0.740 & 0.220 & 0.706 & 0.421 & 0.554
& 0.321 & 0.238 & 0.143 & 0.016 & 0.180 \\

BI + Dolci
& 0.469 & 0.709 & 0.640 & 0.759 & 0.240 & 0.733 & 0.459 & 0.573
& 0.372 & 0.244 & 0.146 & 0.014 & 0.194 \\

Iterative + Dolci
& 0.465 & 0.717 & 0.536 & 0.685 & 0.280 & 0.731 & 0.387 & 0.543
& 0.260 & 0.165 & 0.116 & 0.070 & 0.153 \\

\addlinespace[0.5mm]
\multicolumn{14}{l}{\textbf{\emph{Self-Generated Responses}}} \\

\textit{SGR (Reverse + S.Alpaca)}
& 0.403 & 0.688 & 0.617 & 0.685 & 0.320 & 0.660 & 0.400 & 0.539
& 0.445 & 0.159 & 0.077 & 0.037 & 0.179 \\

\textit{SGR (BI + S.Alpaca)}
& 0.495 & 0.731 & 0.657 & 0.776 & 0.260 & 0.777 & 0.472 & 0.595
& 0.574 & 0.226 & 0.228 & 0.178 & 0.302 \\

\textit{SGR (Reverse + S.Dolci)}
& 0.478 & 0.736 & 0.654 & 0.724 & 0.270 & 0.750 & 0.455 & 0.581
& 0.498 & 0.305 & 0.119 & 0.006 & 0.232 \\

\textit{SGR (BI + S.Dolci)}
& 0.488 & 0.726 & 0.668 & 0.764 & 0.260 & 0.756 & 0.455 & 0.588
& 0.601 & 0.348 & 0.185 & 0.156 & 0.323 \\

\textit{SGR (Iterative + S.Dolci)}
& 0.468 & 0.726 & 0.567 & 0.756 & 0.320 & 0.716 & 0.404 & 0.565
& 0.513 & 0.232 & 0.148 & 0.158 & 0.263 \\

\midrule
\multicolumn{14}{l}{\textbf{Qwen2.5-7B-Instruct}} \\
\midrule
\multicolumn{14}{l}{\textbf{\emph{Open-source data}}} \\

Reverse + Alpaca
& 0.390 & 0.658 & 0.567 & 0.656 & 0.200 & 0.632 & 0.357 & 0.494
& 0.011 & 0.018 & 0.016 & 0.120 & 0.041 \\

BI + Alpaca
& 0.476 & 0.777 & 0.356 & 0.606 & 0.380 & 0.767 & 0.443 & 0.544
& 0.087 & 0.122 & 0.217 & 0.156 & 0.145 \\

Reverse + Dolci
& 0.374 & 0.628 & 0.521 & 0.634 & 0.200 & 0.571 & 0.328 & 0.465
& 0.053 & 0.116 & 0.095 & 0.110 & 0.093 \\

BI + Dolci
& 0.450 & 0.758 & 0.376 & 0.656 & 0.360 & 0.740 & 0.392 & 0.533
& 0.241 & 0.146 & 0.220 & 0.166 & 0.193 \\

Iterative + Dolci
& 0.454 & 0.720 & 0.519 & 0.660 & 0.300 & 0.727 & 0.455 & 0.548
& 0.263 & 0.244 & 0.283 & 0.165 & 0.239 \\

\addlinespace[0.5mm]
\multicolumn{14}{l}{\textbf{\emph{Self-Generated Responses}}} \\

\textit{SGR (Reverse + S.Alpaca)}
& 0.406 & 0.677 & 0.628 & 0.650 & 0.200 & 0.651 & 0.396 & 0.515
& 0.022 & 0.037 & 0.019 & 0.123 & 0.050 \\

\textit{SGR (BI + S.Alpaca)}
& 0.489 & 0.777 & 0.400 & 0.634 & 0.300 & 0.748 & 0.468 & 0.545
& 0.181 & 0.134 & 0.228 & 0.172 & 0.179 \\

\textit{SGR (Reverse + S.Dolci)}
& 0.372 & 0.623 & 0.542 & 0.562 & 0.212 & 0.572 & 0.361 & 0.463
& 0.137 & 0.104 & 0.124 & 0.122 & 0.122 \\

\textit{SGR (BI + S.Dolci)}
& 0.467 & 0.777 & 0.404 & 0.646 & 0.350 & 0.761 & 0.434 & 0.548
& 0.294 & 0.207 & 0.315 & 0.175 & 0.248 \\

\textit{SGR (Iterative + S.Dolci)}
& 0.468 & 0.739 & 0.518 & 0.654 & 0.250 & 0.763 & 0.485 & 0.554
& 0.519 & 0.238 & 0.296 & 0.167 & 0.305 \\

\midrule
\multicolumn{14}{l}{\textbf{Mistralv0.3-7B-Instruct}} \\
\midrule
\multicolumn{14}{l}{\textbf{\emph{Open-source data}}} \\

Reverse + Alpaca
& 0.476 & 0.720 & 0.474 & 0.701 & 0.230 & 0.719 & 0.498 & 0.545
& 0.040 & 0.055 & 0.032 & 0.016 & 0.036 \\

BI + Alpaca
& 0.473 & 0.707 & 0.507 & 0.709 & 0.310 & 0.714 & 0.413 & 0.548
& 0.026 & 0.085 & 0.029 & 0.008 & 0.037 \\

Reverse + Dolci
& 0.453 & 0.696 & 0.520 & 0.744 & 0.260 & 0.695 & 0.374 & 0.535
& 0.173 & 0.110 & 0.063 & 0.018 & 0.091 \\

BI + Dolci
& 0.465 & 0.707 & 0.540 & 0.752 & 0.270 & 0.666 & 0.379 & 0.540
& 0.109 & 0.110 & 0.037 & 0.012 & 0.067 \\

Iterative + Dolci
& 0.461 & 0.739 & 0.422 & 0.673 & 0.300 & 0.681 & 0.387 & 0.523
& 0.135 & 0.085 & 0.034 & 0.016 & 0.068 \\

\addlinespace[0.5mm]
\multicolumn{14}{l}{\textbf{\emph{Self-Generated Responses}}} \\

\textit{SGR (Reverse + S.Alpaca)}
& 0.501 & 0.742 & 0.549 & 0.685 & 0.340 & 0.765 & 0.506 & 0.584
& 0.084 & 0.098 & 0.074 & 0.009 & 0.066 \\

\textit{SGR (BI + S.Alpaca)}
& 0.492 & 0.728 & 0.569 & 0.677 & 0.270 & 0.721 & 0.468 & 0.561
& 0.047 & 0.092 & 0.074 & 0.014 & 0.057 \\

\textit{SGR (Reverse + S.Dolci)}
& 0.492 & 0.720 & 0.572 & 0.717 & 0.330 & 0.748 & 0.472 & 0.579
& 0.194 & 0.183 & 0.114 & 0.016 & 0.127 \\

\textit{SGR (BI + S.Dolci)}
& 0.488 & 0.715 & 0.559 & 0.721 & 0.270 & 0.704 & 0.481 & 0.563
& 0.182 & 0.177 & 0.093 & 0.011 & 0.116 \\

\textit{SGR (Iterative + S.Dolci)}
& 0.481 & 0.726 & 0.442 & 0.661 & 0.300 & 0.725 & 0.409 & 0.535
& 0.106 & 0.140 & 0.064 & 0.007 & 0.079 \\

\bottomrule
\end{tabular}
}
\caption{
Raw performance of the pruned models across classification and generative
benchmarks. SGR is our method with the pruning
metric in parentheses. Reverse, BI, and Iterative (described in Algorithm
\ref{alg:greedy_pruning}) indicate the pruning order, while Alpaca and Dolci
denote the training data source; S.Alpaca and S.Dolci refer to self-generated
variants of the corresponding datasets.
}
\label{tab:raw_joint_full}
\end{table}

\subsection{Pruning at Moderate Ratios}

\label{sec:pruning_moderate}

\begin{figure}[htbp]
  \centering
  \includegraphics[width=0.9\columnwidth]{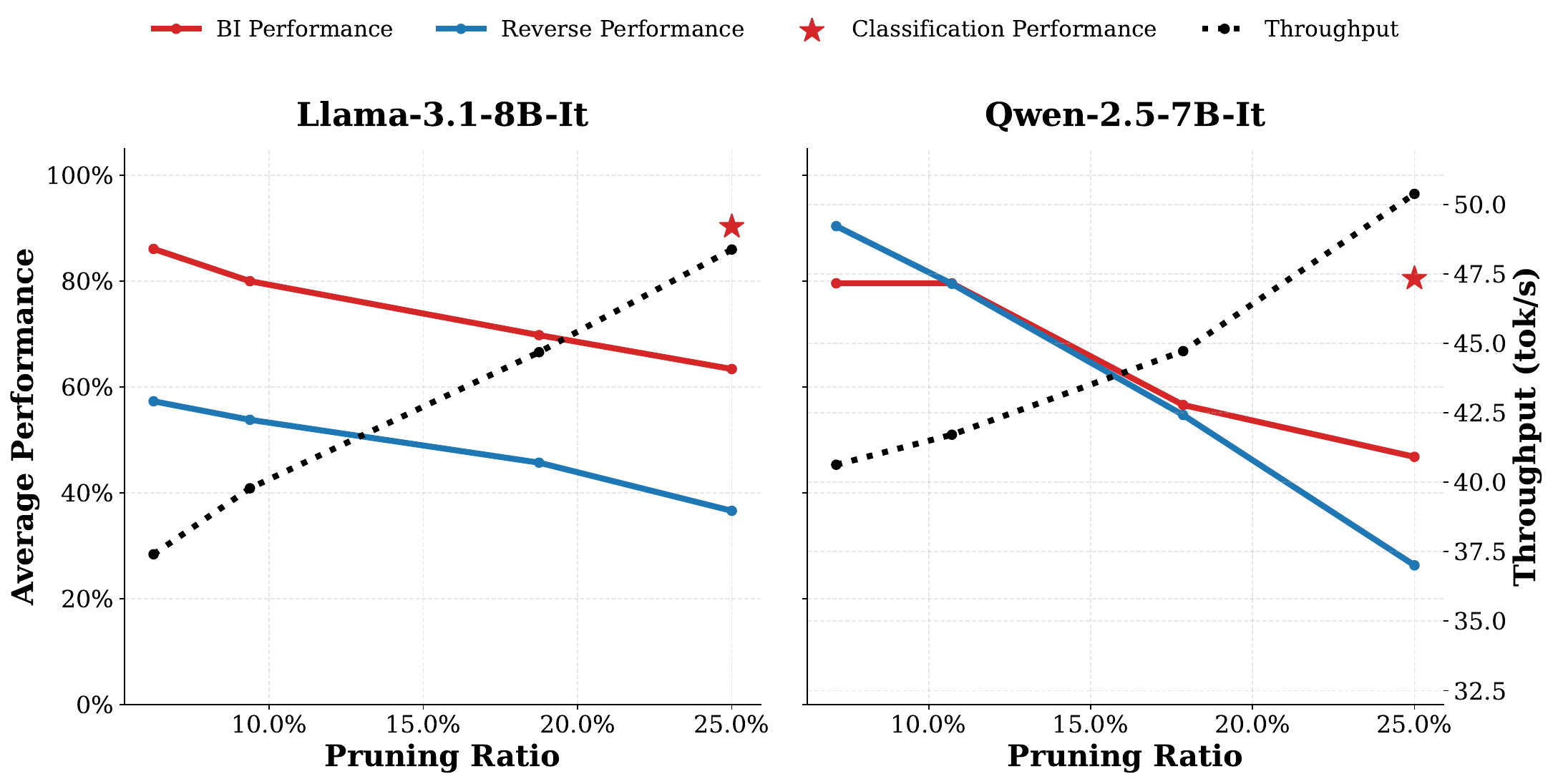}
  \caption{Average accuracy on generative tasks for Qwen and LLaMA across pruning strategies and ratios, alongside model throughput. Throughput (tokens/s) is shown on the secondary axis (dotted line). Even at moderate pruning ratio (25\%), a substantial gap remains in generative vs classification tasks. Additional details are provided in Appendix~\ref{throughput-ratios}.}
  \label{fig:pruning-ratios}
\end{figure}

While prior sections highlight the limitations of aggressive pruning, moderate depth reduction remains appealing in practice due to its simplicity and direct impact on inference latency. Unlike other compression techniques, removing entire transformer blocks yields immediate throughput gains without requiring specialized hardware or kernel-level optimizations.

To better understand this trade-off, we evaluate performance across varying pruning ratios (Figure~\ref{fig:pruning-ratios}; full results in Table~\ref{tab:pruning_limits_recovery}). With only two layers removed, both models retain approximately 85--90\% of their original performance while achieving a \textbf{1.1--1.15}$\times$ increase in throughput. However, performance degrades steadily beyond this point, particularly for generative tasks, making more aggressive pruning less practical when reasoning capabilities are required.

We also observe that pruning strategies optimized for classification do not necessarily generalize to generative reasoning. For example, while Reverse pruning is competitive with Block Influence (BI) on classification tasks for LLaMA, it underperforms on generative benchmarks, reinforcing the earlier observation that classification metrics alone are insufficient for guiding pruning decisions.

Other approaches such as quantization \citep{frantar2022gptq,dettmers2023case,lin2024awq} and sparsification \citep{frantar2023sparsegptmassivelanguagemodels,sun2024simpleeffectivepruningapproach} are effective at reducing memory footprint, but often yield limited throughput gains under realistic settings and may require specialized support. In contrast, layer pruning provides a simple and hardware-agnostic mechanism for improving latency, and can be combined with these techniques for complementary benefits. Our results suggest that such approaches are most effective when applied conservatively, as increasing depth reduction leads to disproportionate degradation in generative reasoning performance.

\subsection{Throughput under Different Pruning Ratios}
\label{throughput-ratios}

To complement Section~\ref{sec:pruning_moderate}, we report decoding throughput under moderate pruning ratios using the Block Influence (BI) strategy. Following prior work (e.g., \citep{frantar2022gptq}), we measure tokens-per-second on WikiText-2 with a fixed context length of 512 tokens, which provides a lightweight proxy for autoregressive generation efficiency.

We group pruning configurations into three regimes based on the proportion of removed layers: approximately 10\%, 18\%, and 25\%. All measurements are conducted under identical hardware and inference settings, and we report throughput, relative speedup, and GPU memory usage.

\begin{table}[t]
\centering
\small
\setlength{\tabcolsep}{6pt}
\begin{tabular}{lccc}
\toprule
\textbf{Pruning Level}
& \textbf{Throughput (tok/s)}
& \textbf{Speedup}
& \textbf{GPU Mem (GB)} \\

\midrule

\multicolumn{4}{l}{\textbf{LLaMA-3.1-8B-Instruct}} \\
\midrule
Base
& 35.73 & 1.00$\times$ & 14.96 \\
$\sim$10\% pruned
& 40.37 & 1.13$\times$ & 13.74 \\
$\sim$18\% pruned
& 43.75 & 1.23$\times$ & 12.52 \\
$\sim$25\% pruned
& 46.30 & 1.30$\times$ & 11.71 \\

\midrule

\multicolumn{4}{l}{\textbf{Qwen2.5-7B-Instruct}} \\
\midrule
Base
& 38.47 & 1.00$\times$ & 14.19 \\
$\sim$10\% pruned
& 41.71 & 1.08$\times$ & 12.88 \\
$\sim$18\% pruned
& 44.72 & 1.16$\times$ & 12.01 \\
$\sim$25\% pruned
& 50.38 & 1.31$\times$ & 11.15 \\

\bottomrule
\end{tabular}
\caption{
Throughput and memory under BI pruning at moderate pruning ratios.
Pruning levels are grouped into approximate percentages of removed layers.
}
\end{table}

We observe a consistent increase in decoding throughput as the pruning ratio increases. For both model families, removing approximately 25\% of layers yields a speedup of around 1.3$\times$, while also reducing GPU memory usage by roughly 20--25\%.

Importantly, we find that these throughput trends closely match those observed during evaluation on generative benchmarks (e.g., GSM8K, HumanEval+, MBPP+, XSUM), indicating that WikiText-2 serves as a reliable proxy for relative efficiency comparisons.

However, as discussed in Section~\ref{sec:pruning_moderate}, these gains come with a noticeable degradation in generative reasoning performance. While moderate pruning (e.g., $\sim$10--18\%) provides a favorable trade-off between efficiency and performance, more aggressive pruning (e.g., $\sim$25\%) leads to disproportionately larger drops in reasoning capability relative to the additional speedup.

\subsection{Pruning at Different Ratios}
\label{pruning-low-ratio-appendix}

\begin{table}[t]
\centering
\small
\setlength{\tabcolsep}{6pt}
\begin{tabular}{lcccc|c}
\toprule
\textbf{Method}
& \textbf{GSM8K}
& \textbf{HumanEval}
& \textbf{MBPP}
& \textbf{XSUM}
& \textbf{Average} \\
\midrule

\multicolumn{6}{l}{\textbf{Qwen}} \\
\midrule

\multicolumn{6}{l}{\textit{2 Layers Pruned}} \\
BI
& 0.878 & 0.692 & 0.709 & 0.906 & 0.796 \\
Iterative
& 0.940 & 0.817 & 0.736 & 0.930 & 0.856 \\
Reverse
& 0.946 & 0.950 & 0.793 & 0.925 & \textbf{0.904} \\

\midrule
\multicolumn{6}{l}{\textit{3 Layers Pruned}} \\
BI
& 0.878 & 0.692 & 0.709 & 0.906 & \textbf{0.796} \\
Iterative
& 0.896 & 0.633 & 0.421 & 0.910 & 0.715 \\
Reverse
& 0.785 & 0.817 & 0.685 & 0.891 & 0.795 \\

\midrule
\multicolumn{6}{l}{\textit{5 Layers Pruned}} \\
BI
& 0.538 & 0.409 & 0.496 & 0.822 & 0.566 \\
Iterative
& 0.819 & 0.658 & 0.458 & 0.853 & \textbf{0.697} \\
Reverse
& 0.422 & 0.476 & 0.455 & 0.837 & 0.547 \\

\midrule
\multicolumn{6}{l}{\textit{7 Layers Pruned}} \\
BI
& 0.329 & 0.283 & 0.398 & 0.861 & 0.468 \\
Iterative
& 0.591 & 0.350 & 0.358 & 0.797 & \textbf{0.524} \\
Reverse
& 0.153 & 0.142 & 0.157 & 0.600 & 0.263 \\

\midrule

\multicolumn{6}{l}{\textbf{LLaMA}} \\
\midrule

\multicolumn{6}{l}{\textit{2 Layers Pruned}} \\
BI
& 0.932 & 0.811 & 0.726 & 0.977 & \textbf{0.861} \\
Iterative
& 0.938 & 0.855 & 0.631 & 0.968 & 0.848 \\
Reverse
& 0.871 & 0.855 & 0.530 & 0.037 & 0.573 \\

\midrule
\multicolumn{6}{l}{\textit{3 Layers Pruned}} \\
BI
& 0.926 & 0.755 & 0.558 & 0.963 & 0.800 \\
Iterative
& 0.946 & 0.800 & 0.603 & 0.958 & \textbf{0.827} \\
Reverse
& 0.893 & 0.733 & 0.502 & 0.023 & 0.538 \\

\midrule
\multicolumn{6}{l}{\textit{6 Layers Pruned}} \\
BI
& 0.818 & 0.655 & 0.530 & 0.788 & \textbf{0.698} \\
Iterative
& 0.806 & 0.566 & 0.497 & 0.894 & 0.691 \\
Reverse
& 0.762 & 0.700 & 0.341 & 0.025 & 0.457 \\

\midrule
\multicolumn{6}{l}{\textit{8 Layers Pruned}} \\
BI
& 0.758 & 0.634 & 0.390 & 0.754 & \textbf{0.634} \\
Iterative
& 0.671 & 0.466 & 0.312 & 0.758 & 0.552 \\
Reverse
& 0.628 & 0.556 & 0.251 & 0.029 & 0.366 \\
\bottomrule
\end{tabular}
\caption{
Retention on generative benchmarks under increasing pruning levels.
We report results on GSM8K, HumanEval, MBPP, and XSUM. Average denotes the mean recovery across benchmarks.
}
\label{tab:pruning_limits_recovery}
\end{table}

\begin{figure}[h]
  \begin{center}
    \centerline{\includegraphics[width=0.5\columnwidth]{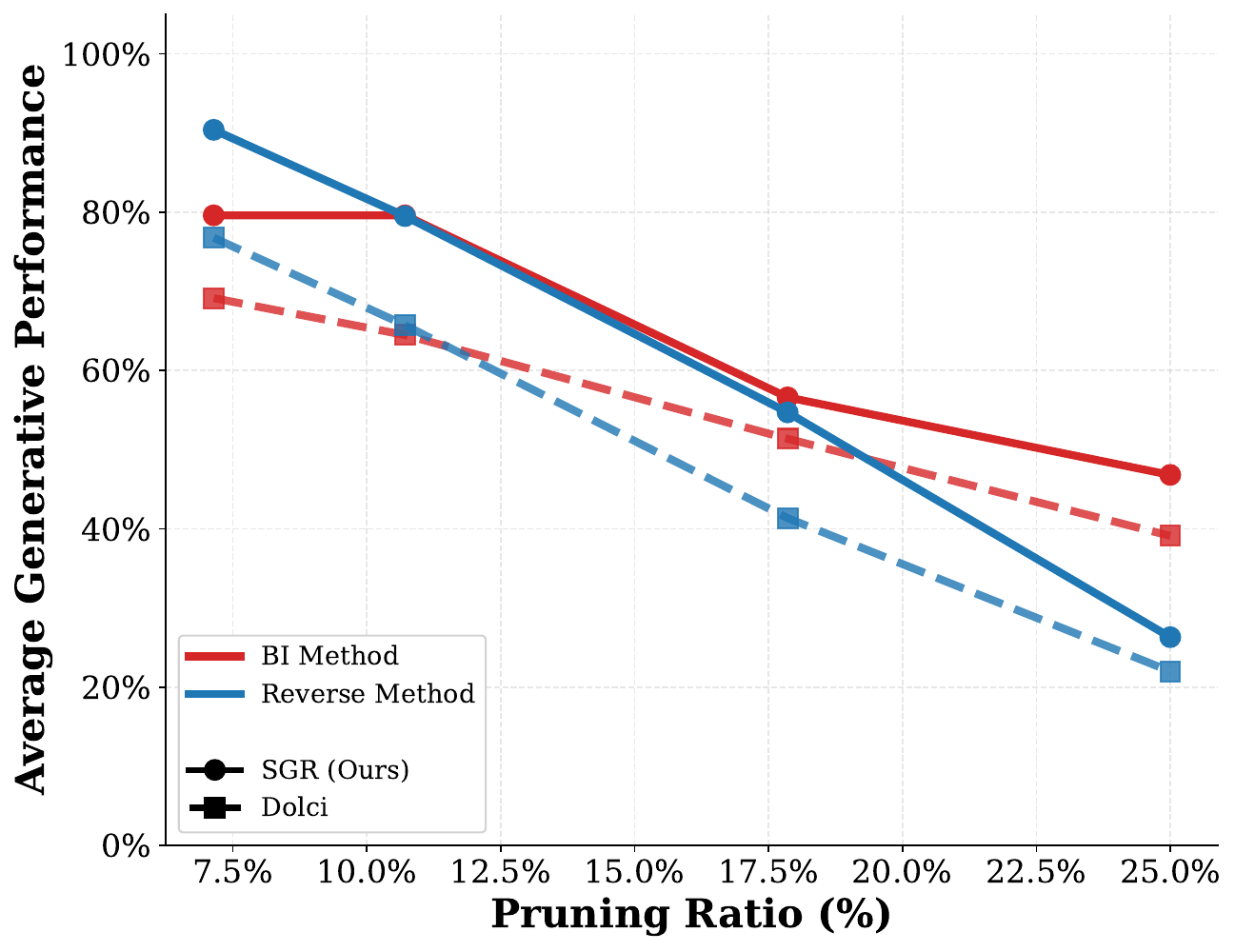}}
    \caption{
      Differences between finetuning with Self-Generated Responses (SGR) vs on Dolci dataset directly for the Qwen Model. At all pruning ratios, SGR is consistently better than the raw dataset.
    }
    \label{fig:qwen_pruning_ratios}
  \end{center}
\end{figure}

% ============================================================
% Additional experiments (from the ARR rebuttal)
% ============================================================

\section{Additional Recovery and Diagnostic Experiments}
\label{app:rebuttal}

This appendix reports experiments added after review. Unless stated otherwise, they use Qwen2.5-7B-Instruct with seven layers (25\%) removed by Block Influence (BI). In the main setting, this model is recovered with QLoRA on self-generated Dolci responses (S.Dolci), following Appendix~\ref{all-finetuning-appendix}. This corresponds to the \textit{SGR (BI + S.Dolci)} row for Qwen in Table~\ref{tab:retention_joint_full}.

% ------------------------------------------------------------
\subsection{Scaling the Recovery Budget}
\label{app:scaling-recovery}

To test whether the recovery gap is an artifact of parameter-efficient tuning or limited data, we recover the same pruned Qwen model with two larger budgets:
(i) \textbf{full finetuning} on the same S.Dolci responses used in the main setting, and
(ii) \textbf{QLoRA on 1M self-generated responses}, which uses about $10\times$ the training data and compute of the main setting.
These runs complement the task-aligned full finetuning on GSM8K (Appendix~\ref{upper-bound}) and the QLoRA vs.\ full-finetuning comparison (Appendix~\ref{qlora_vs_ft}), both of which evaluate only GSM8K. Here we evaluate on the full generative suite.
Table~\ref{tab:scaling-raw} reports raw scores, and Table~\ref{tab:scaling-retention} reports retention relative to the unpruned model.

\begin{table}[h]
\centering
\small
\begin{tabular}{lccccc}
\toprule
\textbf{Recovery setting} & \textbf{GSM8K} & \textbf{HumEval$^{+}$} & \textbf{MBPP$^{+}$} & \textbf{XSUM} & \textbf{Mean} \\
\midrule
QLoRA, S.Dolci (main setting) & 0.294 & 0.207 & 0.315 & 0.175 & 0.248 \\
Full FT, S.Dolci              & 0.415 & 0.220 & 0.310 & 0.158 & 0.276 \\
QLoRA, 1M SGR ($10\times$)    & 0.479 & 0.299 & 0.206 & 0.145 & 0.282 \\
\bottomrule
\end{tabular}
\caption{Raw generative benchmark scores of BI-pruned Qwen2.5-7B-Instruct (seven layers removed) under larger recovery budgets. The first row is taken from Table~\ref{tab:raw_joint_full}.}
\label{tab:scaling-raw}
\end{table}

\begin{table}[h]
\centering
\small
\begin{tabular}{lccccc}
\toprule
\textbf{Recovery setting} & \textbf{GSM8K} & \textbf{HumEval$^{+}$} & \textbf{MBPP$^{+}$} & \textbf{XSUM} & \textbf{Mean} \\
\midrule
QLoRA, S.Dolci (main setting) & 0.329 & 0.283 & 0.398 & 0.861 & 0.468 \\
Full FT, S.Dolci              & 0.464 & 0.300 & 0.391 & 0.781 & 0.484 \\
QLoRA, 1M SGR ($10\times$)    & 0.536 & 0.408 & 0.261 & 0.715 & 0.480 \\
\bottomrule
\end{tabular}
\caption{Performance retention (normalized to the unpruned baseline) for the settings in Table~\ref{tab:scaling-raw}. The first row is taken from Table~\ref{tab:retention_joint_full}.}
\label{tab:scaling-retention}
\end{table}

As Table~\ref{tab:scaling-retention} shows, the larger budgets improve GSM8K retention (0.329 $\rightarrow$ 0.464 and 0.536) and, with 1M-response QLoRA, HumanEval\textsuperscript{+} retention (0.283 $\rightarrow$ 0.408). The gains are not uniform, however. MBPP\textsuperscript{+} drops to 0.261 under 1M-response QLoRA, and XSUM declines in both settings. Mean generative retention therefore barely changes: 0.468 in the main setting, 0.484 with full finetuning, and 0.480 with $10\times$ more data and compute. Together with Section~\ref{classification-result-section} and Appendix~\ref{upper-bound}, this indicates that the recovery gap is not an artifact of QLoRA or of the size of the recovery set.

% ------------------------------------------------------------
\subsection{Single-Token versus Autoregressive Arithmetic}
\label{app:single-vs-ar}

Classification benchmarks are scored by comparing candidate likelihoods, whereas generative benchmarks require multi-token autoregressive decoding, where small per-token shifts can compound (Section~\ref{classification-result-section}). To separate loss of arithmetic ability from this sensitivity, we evaluate Qwen on the arithmetic problems of Section~\ref{arithmetic-error} (dataset details in Appendix~\ref{arithmetic-ablation-details}) in two ways:
(i) \textbf{single-token prediction}, where the model must assign the highest probability to the correct answer digit (the probe of Section~\ref{arithmetic-error}, also used in Figure~\ref{fig:arithmetic-post-recovery}), and
(ii) \textbf{autoregressive generation}, where the model generates a free-form answer that we then parse.
Table~\ref{tab:single-vs-ar} reports results for the unpruned, pruned, and SGR-recovered models.

\begin{table}[h]
\centering
\small
\begin{tabular}{lcc}
\toprule
\textbf{Qwen2.5-7B-Instruct} & \textbf{Single-token} & \textbf{Autoregressive} \\
\midrule
Unpruned            & 96.0 & 80.0 \\
Pruned              & 37.0 & 22.0 \\
Pruned + SGR        & 74.0 & 59.0 \\
\bottomrule
\end{tabular}
\caption{Arithmetic accuracy (\%) under single-token prediction and autoregressive generation.}
\label{tab:single-vs-ar}
\end{table}

As Table~\ref{tab:single-vs-ar} shows, autoregressive generation is less accurate than single-token prediction at every stage, as expected from compounding errors. However, pruning also sharply reduces single-token accuracy (96\% $\rightarrow$ 37\%), and SGR recovers it only partly (74\%). Error accumulation during generation therefore does not fully explain the degradation. Consistent with Section~\ref{arithmetic-recovery}, arithmetic performance declines even when generation is removed.

% ------------------------------------------------------------
\subsection{Controlled Bracket-Completion Probe}
\label{app:bracket-probe}

Our evidence for degraded parenthesis tracking in Sections~\ref{parenthesis-tracking} and~\ref{syntax-recovery} is observational: error categories obtained from execution feedback (Figures~\ref{fig:syntax-error} and~\ref{fig:coding-compression}) and examples such as the one in Appendix~\ref{parenthesis-error}. To complement it with a controlled probe, we built 200 synthetic Python expressions, each missing exactly one final closing delimiter (\texttt{]}, \texttt{)}, or \texttt{\}}). The expressions were generated with GPT-5.6 Sol. The model must produce the correct missing delimiter. Results are shown in Table~\ref{tab:bracket-probe}.

\begin{table}[h]
\centering
\small
\begin{tabular}{lc}
\toprule
\textbf{Qwen2.5-7B-Instruct} & \textbf{Accuracy (\%)} \\
\midrule
Unpruned     & 85.0 \\
Pruned       & 40.0 \\
Pruned + SGR & 49.5 \\
\bottomrule
\end{tabular}
\caption{Accuracy on 200 synthetic bracket-completion examples.}
\label{tab:bracket-probe}
\end{table}

As Table~\ref{tab:bracket-probe} shows, pruning roughly halves delimiter-completion accuracy, and SGR recovers only 9.5 points. Together with the arithmetic probe (Appendix~\ref{app:single-vs-ar}), this gives controlled evidence that delimiter tracking remains impaired after recovery, consistent with the persistent unbalanced-parenthesis errors in Section~\ref{syntax-recovery}.

% ------------------------------------------------------------
\subsection{GSM8K Error Audit}
\label{app:error-audit}

Section~\ref{arithmetic-error} links arithmetic failures to GSM8K performance drops using the single-layer ablations (Figure~\ref{fig:multi-model-ablation}). To estimate more directly how much of the GSM8K degradation comes from arithmetic errors, we randomly sampled 100 incorrect GSM8K outputs each from the unpruned Qwen model and from the pruned + SGR model. GPT-5.6 Sol labeled each error with one of four categories: \emph{logic} (wrong reasoning or setup), \emph{arithmetic only} (correct reasoning but a calculation error, as in Appendix~\ref{arithmetic-mistake-example}), \emph{both}, or \emph{extraction} (correct answer present but not extracted, or a formatting failure). Table~\ref{tab:error-audit} gives the counts.

\begin{table}[h]
\centering
\small
\begin{tabular}{lcccc}
\toprule
\textbf{Qwen2.5-7B-Instruct} & \textbf{Logic} & \textbf{Arithmetic only} & \textbf{Both} & \textbf{Extraction} \\
\midrule
Unpruned     & 67 & 14 & 1 & 18 \\
Pruned + SGR & 66 & 23 & 5 & 6  \\
\bottomrule
\end{tabular}
\caption{Error categories for 100 randomly sampled incorrect GSM8K outputs per model.}
\label{tab:error-audit}
\end{table}

In Table~\ref{tab:error-audit}, the number of logic errors is nearly unchanged (67 vs.\ 66), while arithmetic-only errors rise from 14 to 23, a 64\% relative increase. Errors involving both logic and arithmetic also rise (1 $\rightarrow$ 5). This is consistent with the arithmetic probes (Section~\ref{arithmetic-recovery} and Appendix~\ref{app:single-vs-ar}): after pruning and recovery, a larger share of failures comes from calculation errors rather than reasoning errors. These results describe behavior and do not identify an internal mechanism.

% ------------------------------------------------------------
\subsection{Quality of Self-Generated Training Data}
\label{app:sgr-quality}

SGR keeps all teacher rollouts, correct or not, because its goal is to approximate the unpruned model's output distribution. To check whether incorrect teacher responses could explain the limited recovery, we measured the accuracy of the self-generated GSM8K training responses used in the task-aligned recovery experiments (Section~\ref{upper-bound}; eight responses per training problem). Results are shown in Table~\ref{tab:sgr-quality}.

\begin{table}[h]
\centering
\small
\begin{tabular}{lc}
\toprule
\textbf{Model} & \textbf{Training-data accuracy (\%)} \\
\midrule
Qwen2.5-7B-Instruct      & 94.3 \\
Llama-3.1-8B-Instruct    & 86.0 \\
Mistral-7B-Instruct-v0.3 & 42.6 \\
\bottomrule
\end{tabular}
\caption{Accuracy of the self-generated GSM8K training responses used for task-aligned recovery (Appendix~\ref{upper-bound}).}
\label{tab:sgr-quality}
\end{table}

As Table~\ref{tab:sgr-quality} shows, the Qwen and Llama training data are highly accurate, yet both models remain well below their unpruned baselines after task-aligned full finetuning (Figure~\ref{fig:qlora_vs_sft_paper}). Qwen peaks at 65.2\% vs.\ 89.4\%, and Llama at 51.5\% vs.\ 79.3\%. Mistral's training data is much noisier, which may contribute to its weaker recovery. Incorrect teacher responses therefore cannot by themselves explain the recovery gap.

% ------------------------------------------------------------
\subsection{GSM8K Checkpoint Selection Protocol}
\label{app:gsm8k-protocol}

The task-aligned GSM8K runs (Figures~\ref{fig:qlora_vs_sft_paper} and~\ref{fig:qlora_vs_sft}) use the fixed two-epoch schedule described in Appendix~\ref{upper-bound}. The 10\% response-level validation split (5,978 examples) is used only to monitor training loss. It is not used for early stopping or checkpoint selection. The peak accuracies we report describe the trajectory on the disjoint GSM8K test set, not a checkpoint chosen on that test set.

% TODO (requested by the meta-review; not covered in the rebuttal text):
%   - SGR sampling temperature, response lengths, and response correctness for the main (Dolci/Alpaca) runs
%   - RLVR (Appendix~\ref{rlvr-recovery-appendix}): reward design, optimizer, and key hyperparameters

\end{document}